\documentclass[letterpaper, 10 pt, journal, twoside]{IEEEtran}

\usepackage{cite}
\usepackage{amsmath,amssymb,amsfonts}
\usepackage{algorithmic}
\usepackage{graphicx}
\usepackage{textcomp}
\usepackage{xcolor}
\usepackage{lipsum}
\usepackage{fancyhdr}
\ifCLASSOPTIONcompsoc
\usepackage[caption=false,font=normalsize,labelfont=sf,textfont=sf]{subfig}
\else
\usepackage[caption=false,font=footnotesize]{subfig}
\fi

\def\BibTeX{{\rm B\kern-.05em{\sc i\kern-.025em b}\kern-.08em
    T\kern-.1667em\lower.7ex\hbox{E}\kern-.125emX}}

\begin{document}

\title{Discontinuous and Smooth Depth Completion with Binary Anisotropic Diffusion Tensor}



\author{Yasuhiro Yao$^{1,2}$, Menandro Roxas$^{1}$, Ryoichi Ishikawa$^{1}$, Shingo Ando$^{2}$, Jun Shimamura$^{2}$, and Takeshi Oishi$^{1}$%
\thanks{Manuscript received: February, 24, 2020; Revised May, 20, 2020; Accepted June, 16, 2020.}
\thanks{This paper was recommended for publication by Editor Eric Marchand upon evaluation of the Associate Editor and Reviewers' comments. 
}
\thanks{$^{1}$Yasuhiro Yao, Menandro Roxas, Ryoichi Ishikawa, and Takeshi Oishi are with Institute of Industrial Science, the University of Tokyo, Japan
{\tt\small yao@cvl.iis.u-tokyo.ac.jp}}%
\thanks{$^{2} $Yasuhiro Yao, Shingo Ando, and Jun Shimamura are with Media Intelligence Laboratories, Nippon Telegraph and Telephone Corporation, Japan
{\tt\small yasuhiro.yao.tc@hco.ntt.co.jp}}%
\thanks{Digital Object Identifier (DOI): see top of this page.}
}

\maketitle

\thispagestyle{fancy}
\fancyhf{}
\chead{
\scriptsize
\copyright 2020 IEEE.
Personal use of this material is permitted.
Permission from IEEE must be obtained for all other uses, in any current or future media, including reprinting/republishing this material for advertising or promotional purposes, creating new collective works, for resale or redistribution to servers or lists, or reuse of any copyrighted component of this work in other works.
}

\renewcommand{\headrulewidth}{0pt} 

\begin{abstract}
We propose an unsupervised real-time dense depth completion from a sparse depth map guided by a single image.
Our method generates a smooth depth map while preserving discontinuity between different objects.
Our key idea is a Binary Anisotropic Diffusion Tensor (B-ADT) which can completely eliminate smoothness constraint at intended positions and directions by applying it to variational regularization.
We also propose an Image-guided Nearest Neighbor Search (IGNNS) to derive a piecewise constant depth map which is used for B-ADT derivation and in the data term of the variational energy.
Our experiments show that our method can outperform previous unsupervised and semi-supervised depth completion methods in terms of accuracy.
Moreover, since our resulting depth map preserves the discontinuity between objects, the result can be converted to a visually plausible point cloud.
This is remarkable since previous methods generate unnatural surface-like artifacts between discontinuous objects.
\end{abstract}


\section{Introduction}

\IEEEPARstart{L}{ong}-range and real-time depth estimation can be achieved using Light Detection and Ranging (LIDAR) sensors.
They have longer measurable range compared to depth cameras and hence commonly used for outdoor 3D scanning.
But LIDAR can only give sparse measurements in real time (e.g. 10Hz Velodyne sensors \cite{geiger2013vision}) due to the limited number of lasers in its array.

One way to address the sparsity of the depth data is through depth completion.
It is a technique that takes a sparse depth map and other related information, such as RGB images, as inputs and infers a dense depth map.

Although there are supervised depth completion methods in the literature, we focus on unsupervised methods, among which variational methods have been successful.
In \cite{ferstl2013image}, an image-guided Anisotropic Diffusion Tensor (ADT) is applied to the Total Generalized Variation (TGV) regularization which enabled the generation of high-resolution depth images that, although smooth, preserves object boundaries to some extent.

\begin{figure}[t]
\subfloat[Input image]{\includegraphics[width=43mm]{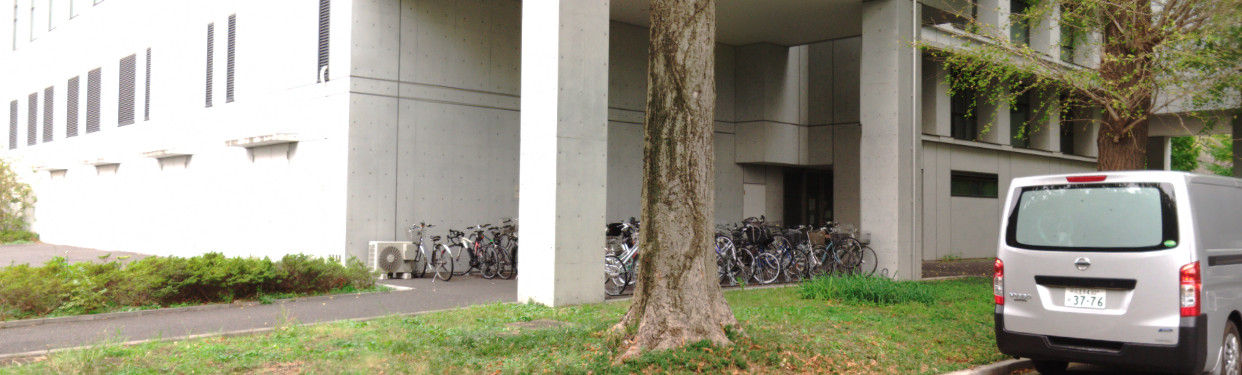}}
\hfil
\subfloat[Input LIDAR]{\includegraphics[width=43mm]{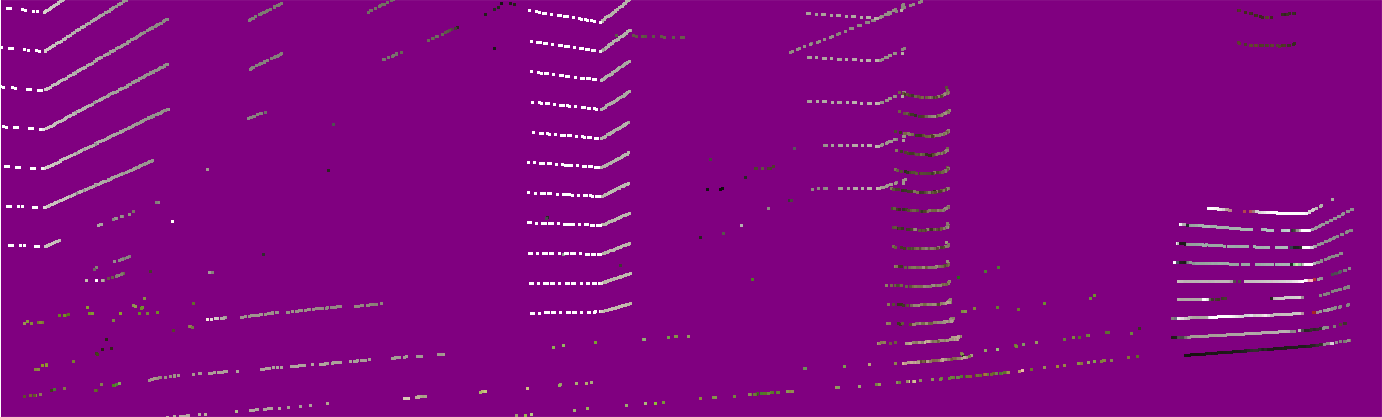}}
\vspace{-8pt}
\subfloat[Result by \cite{ferstl2013image}]{\includegraphics[width=43mm]{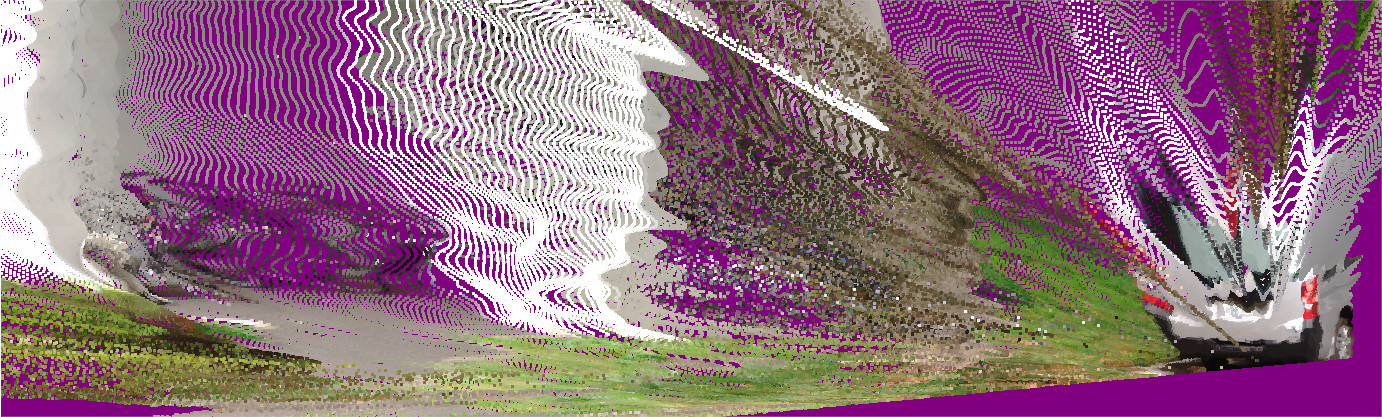}}
\hfil
\subfloat[Our result]{\includegraphics[width=43mm]{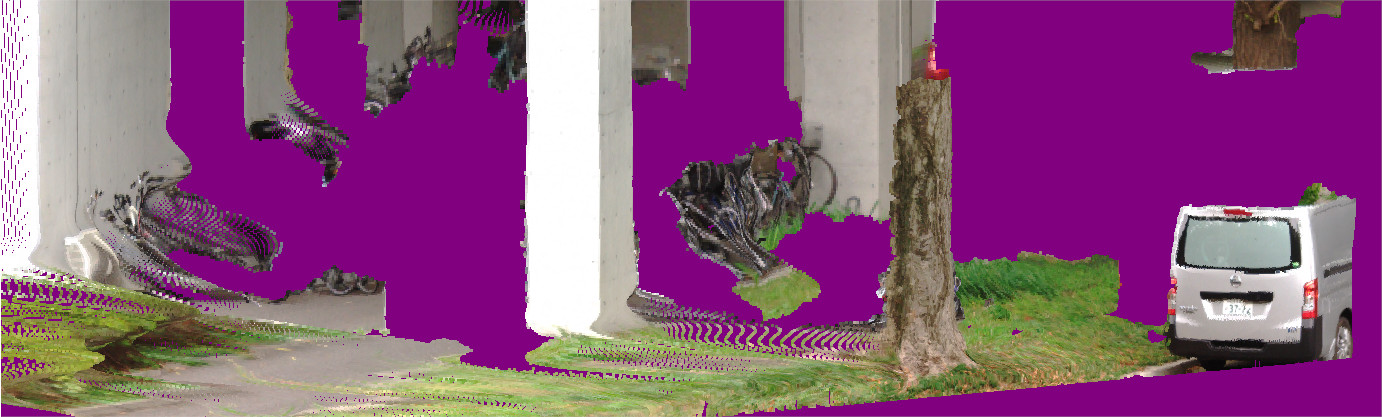}}
\caption{
Depth completion results in point cloud representation.
(c) Multiple objects are connected by points forming surface-like artifacts.
This is because the method imposes smoothness cost across the boundaries of the object.
(d) Our method successfully separates multiple objects because B-ADT enables us to eliminate smoothness cost across the boundaries.
}
\label{fig_abst}
\end{figure}

A disadvantage of the method in \cite{ferstl2013image} is that the depth transition between objects is still continuous.
The effect can be seen as points spreading between discontinuous objects in point cloud representation (Fig. \ref{fig_abst}).
Such points are not plausible because they form surface-like artifacts at the location where no object exists.
They also degrade the depth completion accuracy.

To overcome this, we propose a depth completion method that outputs a dense depth map that is discontinuous at occlusion boundaries and smooth elsewhere.
We introduce a Binary Anisotropic Diffusion Tensor (B-ADT) and an Image-guided Nearest Neighbor Search (IGNNS).
B-ADT enables us to incorporate boundary direction-aware discontinuity in variational methods.
IGNNS enables us to jointly derive the occlusion boundaries, which are used for B-ADT derivation, from an image and a sparse depth map.
IGNNS also serves to give a piecewise constant depth map to be used in the data term of the variational energy.
We show in our experiments the advantages of depth completion with B-ADT and IGNNS against previous and baseline methods.
With our implementation, we were able to achieve real-time processing using modern GPUs.

In summary, our contributions are as follows.
\begin{itemize}
\item[$\bullet$] We propose B-ADT applicable to variational regularization in image processing for eliminating the smoothness constraint at intended pixels and directions.
\item[$\bullet$] We propose a real-time fully unsupervised depth completion method guided by a single image with B-ADT and IGNNS. The proposed method generates depth maps with apparent occlusion boundaries.
\item[$\bullet$] We show in our experiments that our method can outperform previous unsupervised and semi-supervised depth completion methods in terms of accuracy. We also show that our result is suitable for point cloud representation since it preserves discontinuity between objects. Additionally, we provide the parameter sensitivity evaluation of our method.
\end{itemize}

\section{Related Works}
In this section, we discuss the scopes and limitations of existing depth completion methods including supervised and unsupervised methods.

\subsection{Supervised Methods}
Supervised methods can be classified into fully supervised \cite{uhrig2017sparsity, huang2019hms} and semi-supervised \cite{ma2019self, schneider2016semantically, hirata2019real,  wong2020unsupervised}.
Fully supervised methods require ground truth dense depth maps for training. 
However, such dense depth maps are not generally available since it requires the integration of multiple sensors to produce as in \cite{geiger2013vision}.
On the other hand, semi-supervised methods do not require the same training overhead.
Ma et al. \cite{ma2019self} and Wong et al. \cite{wong2020unsupervised} proposed methods with Deep Neural Networks (DNN) which can be self-supervised using the monocular camera frames and sparse depth maps from LIDAR with motion.
However, the method does not apply to scenes without camera and LIDAR motion.
Schneider et al. proposed a method that pre-trained semantic segmentation for guided upsampling of sparse depth maps \cite{schneider2016semantically}, and Hirata et al. used pre-trained semantic segmentation and motion stereo for a similar purpose \cite{hirata2019real}.
In principle, supervised methods apply only to scenes of the same domain as the training. 

\subsection{Unsupervised Methods}
Compared with supervised methods, unsupervised methods are more applicable in situations where a large amount of data for training is not available.
The earliest unsupervised depth completion methods are based on interpolation.
Kopf et al. \cite{kopf2007joint} proposed a method to interpolate low-resolution depth values based on the joint distance of color and space in the high-resolution image.
More recently Ku et al. proposed a method by combining handcrafted classical image processing algorithms \cite{ku2018defense}\footnote{This is the highest-ranked unsupervised method in KITTI depth completion benchmark (http://www.cvlibs.net/datasets/kitti/eval\_depth.php) excluding semi-supervised methods at the time of writing.}.

A disadvantage of the interpolation method is that it introduces undesirable smoothness to the result.
To address this problem, energy minimization methods have been studied.
Diebel and Thrun performed an upsampling using a Markov Random Field (MRF) formulation \cite{diebel2006application}, where the smoothness term is weighted according to texture derivatives.
However, the MRF enhanced method results in surface flattening.
To accommodate this, Ferstl et al. formalized depth completion into ADT-aided TGV regularized energy minimization \cite{ferstl2013image}.
Their method was successfully used to smooth and optimize depth maps in more recent methods \cite{chen2016transforming, hirata2019real}.

Nevertheless, since ADT does not eliminate smoothness along occlusion boundaries, it produces surface-like artifacts between foreground and background objects.
This problem becomes more obvious when we represent the depth as a 3D point cloud.

\begin{figure}[tb]
\centering
\includegraphics[width=86mm,]{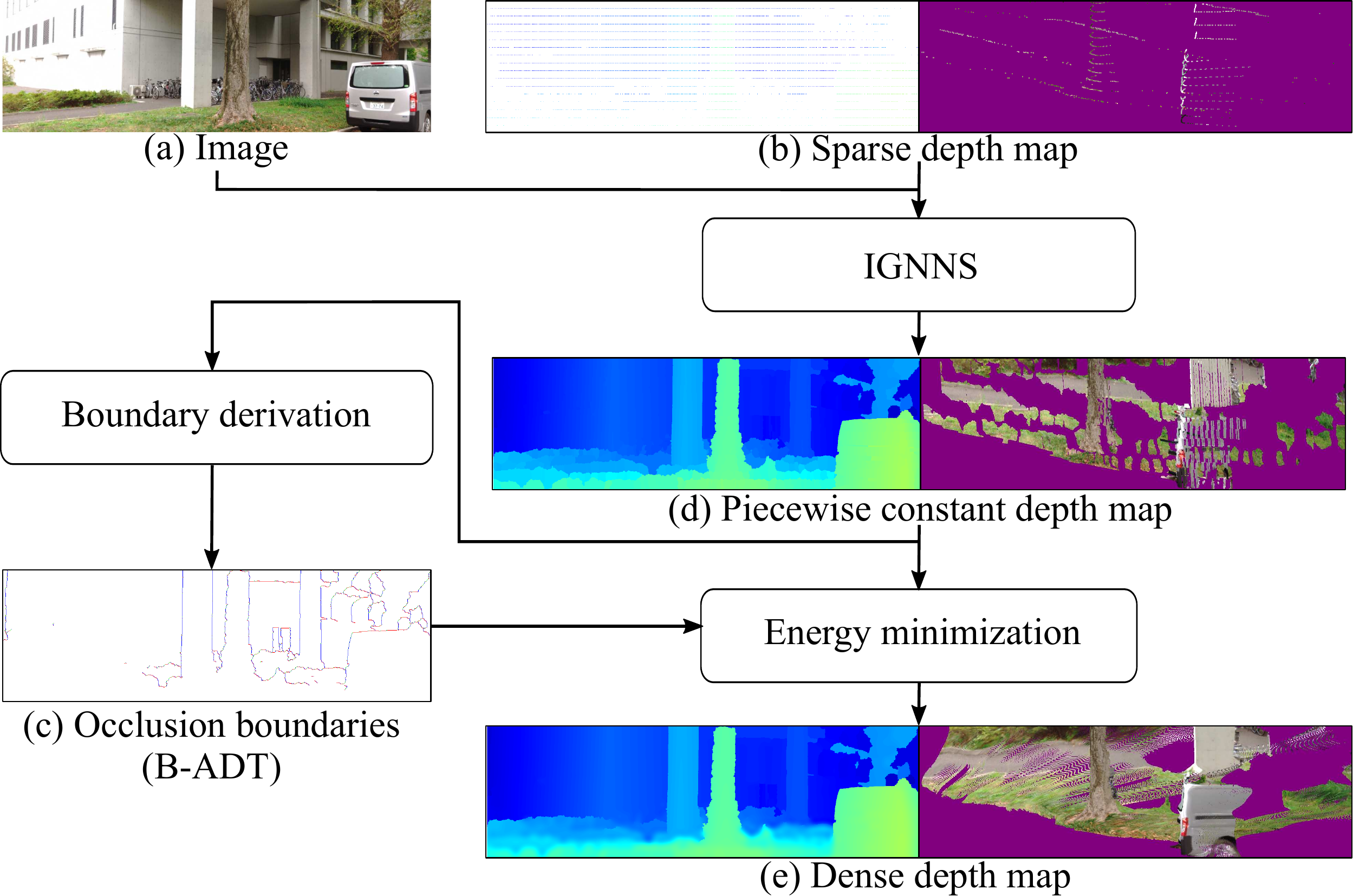}
\caption{
Flow chart of our depth completion.
(b), (d), (e) Left: depth maps, right: point clouds from a side viewpoint.
(c) Colors indicate, white: no occlusion boundary, blue: vertical occlusion boundary, red: horizontal occlusion boundary, green: vertical and horizontal occlusion boundary.
(d) In the piecewise constant depth map, all surfaces are parallel to the image plane.
(e) Our energy minimization smooths depth within an object while preserving discontinuity at object boundaries.
}
\label{fig_flow}
\end{figure}

\section{Our Method}
\label{method}
Our depth completion is composed of IGNNS, boundary (B-ADT) derivation, and the energy minimization (Fig. \ref{fig_flow}).
The core of our method is B-ADT.
We first review ADT and its application as preliminary in Section \ref{related_variation}.
Then, we introduce B-ADT in Section \ref{sec_badt}, IGNNS in Section \ref{sec_nn}, our boundary derivation in Section \ref{sec_boundary}, and our energy minimization in Section \ref{sec_energy}.

We formalized the depth completion problem as follows.
Given an image $I:\Omega \to \mathbb{R}$ and a sparse depth map $d:\Omega \to \mathbb{R}$ where $\Omega \subset \mathbb{R}^2$ is the image domain, our goal is to find the upsampled depth map $u:\Omega \to \mathbb{R}$. 
In our implementation, $I$ is a gray-scale image normalized to the range $[0,1]$.

\subsection{Preliminary: Anisotropic Diffusion Tensor}
\label{related_variation}
ADT was proposed by Werlberger et al. in \cite{werlberger2009anisotropic} for optical flow estimation and applied to other problems such as stereo matching \cite{kuschk2013fast} and depth completion \cite{ferstl2013image}.
ADT serves the purpose of an anisotropic weighting of the regularizer based on the image gradient. 
It enforces low regularization smoothness along image edges and high smoothness in homogeneous image regions.

We denote ADT by $G$ following  \cite{kuschk2013fast}.
ADT is a pixel-wise tensor derived based on the gradient of the input image.
Using an image $I$ and normalized direction of the image gradient ${\bf n} = \frac{\nabla I}{\left| \nabla I \right|}$, pixel-wise ADT $G \in \mathbb{R}^{2 \times 2}$ is derived as follows. 
\begin{equation}
\label{eq_adt}
G = \exp{ \left(-a \left| \nabla I \right|^b \right)} {\bf n} {\bf n}^{\rm T} + {\bf n}^{\bot}{\bf n}^{\bot {\rm T}}
\end{equation}
Here, ${\bf n}^{\bot}$ is the normal vector to the gradient and the scalars $a$ and $b$ are hyper parameters to adjust the magnitude and the sharpness of the tensor.
Note that $G$ is quadratic to ${\bf n}^{\bot}$ in Eq. (\ref{eq_adt}), and hence is not dependent on the choice of normal vector ${\bf n}^{\bot}$ out of two.

ADT has been used to anisotropically weight the TGV regularization first proposed by Bredies et al. in \cite{bredies2010total}.
TGV is composed of polynomials of arbitrary order, which allows us to reconstruct piecewise polynomial functions.

In \cite{ferstl2013image}, the total energy for depth completion is defined as the combination of the data term $C(u)$ and the second order TGV term $R(u,{\bf v})$ with ADT.
With the upsampled depth map $u : \Omega \to \mathbb{R}$ and the relaxation variable ${\bf v} : \Omega \to \mathbb{R}^2$, this energy is defined as: 
\begin{equation}
\label{eq_min_eng}
\int_{\Omega} C(u) + R(u, {\bf v}) d {\bf x}
\end{equation}
where, the data term $C(u)$ and TGV term $R(u, {\bf v})$ are expressed as: 
\begin{align}
\label{eq_data_term}
C(u) = & \lambda_d w \left|  u - d \right|^2 \\
\label{eq_tgv_term}
R(u, {\bf v}) = &  \lambda_s \left|G \left(\nabla u - {\bf v}\right)\right| + \lambda_a \left|\nabla {\bf v} \right| 
\end{align}
Here, $w \in \mathbb{R}$ is the pixel-wise weight for the data term, and hyper parameters $\lambda_s \in \mathbb{R}$, $\lambda_a \in \mathbb{R}$, and $\lambda_d \in \mathbb{R}$ are weights for each of the energy terms.
 $d$ and $w$ are zero at pixels where the depth is not captured.
 
The energy of Eq. (\ref{eq_min_eng}) is convex and can be optimized to a global minimum through the primal dual algorithm \cite{pock2011tgv}.
The energy minimization allows $\nabla u$ to be smooth on $\Omega$ through the relaxation variable ${\bf v}$, while constraining the value $u$ around $d$.
The upsampled depth map is derived as the minimizer of the total energy.

\subsection{Binary Anisotropic Diffusion Tensor}
\label{sec_badt}
Although ADT can control the magnitude of smoothness, it is not sufficient to discretize multiple objects in a scene.
ADT is continuous and still applies weak smoothness constraint at the occlusion boundaries.
The effect is visually apparent in point cloud data as surface-like artifacts connecting discrete objects.

We propose B-ADT to overcome this problem.
B-ADT is an extension of ADT that can eliminate smoothness constraint anisotropically.
As the name suggests, B-ADT is composed of only $0$ and $1$.

B-ADT is derived as extremes of ADT.
We consider two extremes where the magnitude of the image gradient $|\nabla I|$ is either extremely small or extremely large.
More importantly, when $|\nabla I|$ is extremely large, it can totally eliminate the smoothness constraint in the TGV term.
Following the definition of ADT in Eq. (\ref{eq_adt}), we can derive the extremes as follows.
\begin{align}
\label{eq_adt_0}
\lim_{\left|\nabla I\right| \to + 0} G =& {\bf n} {\bf n}^{\rm T} + {\bf n}^{\bot}{\bf n}^{\bot {\rm T}} = 
\left( \begin{array}{cc}
1 & 0 \\
0 & 1 \end{array}\right)
\\
\lim_{\left|\nabla I\right| \to +\infty} G =& {\bf n}^{\bot}{\bf n}^{\bot {\rm T}}
\label{eq_adt_inf}
\end{align}

We use these extremes as B-ADT according to the locations and directions of occlusion boundaries.
Occlusion boundaries are between objects which are not connected to each other and the depth is discontinuous across them.
Note, on the other hand, that the depth is smooth by crossing connected boundaries where different objects are in contact.
We consider the extreme of ADT with $\left|\nabla I \right| \to +\infty$ on occlusion boundaries, and $\left|\nabla I \right| \to +0$ otherwise.

For each pixel, the value of B-ADT is decided based on two conditions, ``$A$: the pixel is on a vertical occlusion boundary'' and ``$B$: the pixel is on a horizontal occlusion boundary''.
Here, a vertical occlusion boundary is a vertical line segment across which the depth is discontinuous in the horizontal direction.
Accordingly, a horizontal occlusion boundary is a horizontal line segment across which the depth is discontinuous in the vertical direction.

\begin{figure}[t]
\centering
\subfloat[Path]{\includegraphics[width=18mm]{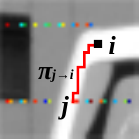}
}
\hfil
\subfloat[Voronoi cells]{\includegraphics[width=18mm]{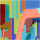}
}
\hfil
\subfloat[Depth map]{\includegraphics[width=18mm]{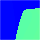}
}
\hfil
\subfloat[Boundary]{\includegraphics[width=18mm]{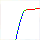}
}
\caption{
Illustration of IGNNS and the boundary derivation.
(a) Pixels with the LIDAR depth are indicated with colors.
The path cost is the weighted sum of the length of the path and accumulation of the norm of the image gradient along the path.
(b) Pixels masked by the same color have the same nearest neighbor.
(c) The piecewise constant depth map is filled with depth value at the nearest neighbor of each pixel.
(d) The occlusion boundaries are derived by thresholding the norm of the gradient of the piecewise constant depth map.
The color scheme in the figure is the same as Fig. \ref{fig_flow} (c)
}
\label{fig_ignns}
\end{figure}

Because the image is defined on 2D grid, every pixel is classified as one of 4 types: ``neither $A$ nor $B$ ($\lnot{A} \land \lnot{B}$)'', ``$A$ but not $B$ ($A \land \lnot{B}$)'', ``not $A$ but $B$ ($\lnot{A} \land B$)'', and ``$A$ and $B$ ($A \land B$)''.
In Fig. \ref{fig_flow} (c) and Fig \ref{fig_ignns} (d), we denote the pixels by $\lnot{A} \land \lnot{B}$ as white, $A \land \lnot{B}$ as blue, $\lnot{A} \land B$ as red, and $A \land B$ as green.
If we let ${\bf n}$ to be defined as ${\bf e}_x$ on vertical and ${\bf e}_y$ on horizontal occlusion boundaries, we can write the B-ADT $\bar{G}$ for each condition as follows.
\begin{align}
\bar{G}_{\lnot{A} \land \lnot{B}} & =  \lim_{\left|\nabla I\right| \to + 0} G =
\left( \begin{array}{cc}
1 & 0 \\
0 & 1 \end{array}\right) \\
\bar{G}_{A \land \lnot{B} } & = \lim_{\left|\nabla I\right| \to +\infty, {\bf n} = {\bf e}_x} G = 
\left( \begin{array}{cc}
0 & 0 \\
0 & 1 \end{array}\right) \\
\bar{G}_{\lnot{A} \land B} & =  \lim_{\left|\nabla I\right| \to +\infty, {\bf n} = {\bf e}_y} G = 
\left( \begin{array}{cc}
1 & 0 \\
0 & 0 \end{array}\right).
\end{align}
In case of $A \land B$, we set B-ADT as product of $\bar{G}_{A \land \lnot{B} }$ and $\bar{G}_{\lnot{A} \land B }$, so that it applies discontinuity in both directions. As a result, we have:
\begin{equation}
\label{eq_A_and_B}
\bar{G}_{A \land B} = \bar{G}_{A \land \lnot{B} } \bar{G}_{\lnot{A} \land B} =
\left( \begin{array}{cc}
0 & 0 \\
0 & 0 \end{array}\right)
\end{equation}

We can verify the effects of B-ADT by applying it to the TGV term in Eq. (\ref{eq_tgv_term}).
By substituting ADT $G$ with B-ADT $\bar{G}$ in Eq. (\ref{eq_tgv_term}), we have our TGV term $\bar{R}(u, {\bf v})$, with denoting ${\bf v} = \left(v_x, v_y \right)^T$, as:
\begin{equation}
\bar{R}(u, {\bf v}) = \left\{
\begin{array}{ll}
    \lambda_s \left| \nabla u - {\bf v}\right| + \lambda_a \left|\nabla {\bf v} \right| & {\rm if}\hspace{2mm} \lnot{A} \land \lnot{B}\\
    \lambda_s \left| \frac{\partial u}{\partial y} - v_y\right| + \lambda_a \left|\nabla {\bf v} \right| & {\rm if} \hspace{2mm} A \land \lnot{B} \\
    \lambda_s \left| \frac{\partial u}{\partial x} - v_x\right| + \lambda_a \left|\nabla {\bf v} \right| & {\rm if} \hspace{2mm} \lnot{A} \land B \\
    \lambda_a \left|\nabla {\bf v} \right| & {\rm if}\hspace{2mm} A \land B \\
\end{array}
\right.
\end{equation}
From the above equations, we can see that there is no smoothness cost on depth across occlusion boundaries.
For example, $x$ directional derivative of depth $u$ does not contribute to the term at all on vertical boundaries.
Hence, an extremely large value of a gradient of $u$ can be taken across the boundaries.
As a consequence, discontinuous changes are allowed across the boundaries with B-ADT.

\subsection{Image-guided Nearest Neighbor Search}
\label{sec_nn}
To derive B-ADT, we need occlusion boundaries well correlated to the depth maps.
For that, we perform IGNNS which searches the shortest path in terms of the accumulation of the image gradient.
IGNNS is based on the observation that two points tend to be on the same object if there is a path between them which does not cross a lot of edges on the image.
The result of IGNNS is also used in our depth completion energy as we see in Section \ref{sec_energy}.
We illustrate IGNNS in Fig. \ref{fig_ignns} (a) - (c).

IGNNS outputs a piecewise constant depth map from an input sparse depth map and an image.
Let $D$ be the set of pixels where the input sparse depth map $d$ has a value and ${\bf i}_\ast \in D$ as the nearest neighbor of ${\bf i} \in \Omega$.
Considering the nearest neighbors of every pixel in the image domain, we get a piecewise constant dense depth map $\bar{d}:\Omega \to \mathbb{R}$ as
\begin {equation}
\label{eq_piecewise}
\bar{d}({\bf i}) = d({\bf i}_\ast).
\end {equation}

IGNNS is a way to derive ${\bf i}_\ast$ with guidance of the input image $I$.
Let $\pi_{{\bf j} \to {\bf i}}$ be a path from pixel ${\bf j}$ to pixel ${\bf i}$ on the image grid.
Here, $\pi_{{\bf j} \to {\bf i}}$ is expressed as set of pixels on the path.
In IGNNS, we search the nearest neighbor ${\bf i}_\ast$ from ${\bf i}$ as:

\begin {equation}
\newcommand{\argmin}{\mathop{\rm argmin}\limits}
\label{eq_nearest}
{\bf i}_\ast = \argmin_{\bf j \in D} \min_{\pi_{{\bf j} \to {\bf i}}} \{
    \sum_{{\bf k} \in \pi_{{\bf j} \to {\bf i}}} \| \nabla I ({\bf k}) \|^2 + c|\pi_{{\bf j} \to {\bf i}}|
    \}
\end {equation}
where the hyper parameter $c$ is the constant cost of the unit path length and $|\pi_{{\bf j} \to {\bf i}}|$ is the number of pixels in $\pi_{{\bf j} \to {\bf i}}$.

With the nearest neighbor ${\bf i}_\ast$ of Eq (\ref{eq_nearest}), we derive the piecewise constant depth map $\bar{d}$ using Eq. (\ref{eq_piecewise}).

\subsection{Boundary derivation}
\label{sec_boundary}
We derive the occlusion boundaries based on IGNNS result $\bar{d}$.
With pre-defined threshold $t$, we decide a pixel ${\bf x}$ to be on a vertical occlusion boundary if $|\frac {\partial \bar{d}({\bf x})}{\partial x}| > t$ and to be on a horizontal occlusion boundary if $|\frac {\partial \bar{d}({\bf x})}{\partial y}| > t$ (Fig. \ref{fig_ignns} (d)).

This thresholding can generate false occlusion boundaries especially on the ground since it is often a large plane parallel to the view direction and has a wide range of depth.
Hence, we filter out occlusion boundaries detected on the ground.
We can generate efficient ground labels from an input sparse depth map as follows.

We convert the input sparse depth map to a point cloud and apply RANSAC plane model segmentation to it.
Then, we label points in and below the detected plane as ground.
Finally, we label each pixel by the label of its IGNNS nearest neighbor.
This process has two hyper parameters regarding RANSAC plane segmentation: the number of iterations $N_{\rm ransac}$ and distance threshold $t_{\rm ransac}$.
Our implementation uses Point Cloud Library \cite{radu20113d} for the plane segmentation. 

In comparison to conventional boundary detection methods relying only on an image as \cite{murasaki2014occlusion}, our method jointly uses an image and a depth map.
Hence the boundaries are well correlated to depth maps and suitable for B-ADT derivation.

\subsection{Energy Minimization}
\label{sec_energy}
We derive the upsampled depth maps by minimizing the energy with B-ADT weighted TGV regularization.
Our energy minimization follows that of ADT aided depth completion \cite{ferstl2013image} with two major differences.
First, we apply B-ADT instead of ADT. Second, we use a densified depth map from raw LIDAR measurements, resulting from IGNNS, for the data term instead of the depth camera as used in \cite{ferstl2013image}.
We used a densified data $\bar{d}$ because we found, through our experiment, that the variational method was very difficult to stably converge using sparse data (see Appendix \ref{ap_conv}).
Moreover, we use inverse depth $\bar{d}^{-1}$ instead of depth $\bar{d}$ to balance the contribution of near and far depth as was done in \cite{newcombe2011dtam}.
To further help the convergence, we scale the inverse depth so that the maximum of $\bar{d}^{-1}$ is one.

We define our data term and TGV term for the energy as:
\begin{align}
\bar{C}(u) &=   \lambda_d w \left|  u - \bar{d}^{-1} \right|^2  \\
\label{eq_our_tgv_term}
\bar{R}(u, {\bf v}) &=  \lambda_s \left|\bar{G} \left(\nabla u - {\bf v}\right)\right| + \lambda_a \left|\nabla {\bf v} \right| 
\end{align}
The depth completion becomes the following energy minimization problem.
\begin{equation}
    \min_{u,{\bf v}}  \int_{\Omega} \bar{C}(u) + \bar{R}(u, {\bf v}) d{\bf x}
\end{equation}
We minimize the above energy using the primal dual algorithm \cite{pock2011tgv}.
First, we discretize the energy and conduct the Legendre-Fenchel transform of TGV term, while introducing dual variables ${\bf p}: \Omega \to \mathbb{R}^2$ and ${\bf q}: \Omega \to \mathbb{R}^4$.
Then $\int \bar{R}(u, {\bf v}) d {\bf x}$ becomes as:
\begin{equation}
\label{eq_lf}
\max_{\|{\bf p}\|_\infty \leq \lambda_s} \langle \bar{G} 
\left(\nabla u - {\bf v}\right) , {\bf p} \rangle +  \max_{\|{\bf q}\|_\infty \leq \lambda_a} \langle \nabla {\bf v}, {\bf q} \rangle.
\end{equation}
Then we apply gradient descent on $u$, ${\bf v}$ and gradient ascent on ${\bf p}$, ${\bf q}$ to derive the primal dual iterations for the energy minimization as
\begin{align}
\label{eq_it_p}
{\bf p}^{n+1} =& \Pi_P \left({\bf p}^n + \tau_p \bar{G}  \left(\nabla \hat{u}^n - \hat{\bf v}^n \right) \right)  \\
\label{eq_it_q}
{\bf q}^{n+1} =& \Pi_Q \left({\bf q}^n + \tau_q \nabla \hat{\bf v}^n \right) \\
\label{eq_it_u}
{u}^{n+1} =& \frac{{u}^n + \tau_u \left( {\rm div}  \bar{G} {\bf p}^{n+1}  +  \lambda_d w \bar{d}^{-1} \right)} {1 + \tau_u \lambda_d w} \\
\label{eq_it_v}
{\bf v}^{n+1} =& {\bf v}^n + \tau_v \left( {\bf p}^{n+1} + {\rm div} {\bf q}^{n+1} \right) \\
\hat{u}^{n+1} =& 2 {u}^{n+1} - {u}^n \\
\hat{{\bf v}}^{n+1} =& 2 {\bf v}^{n+1} - {\bf v}^n , 
\end{align}
where hyper parameters $\tau_p$, $\tau_q$, $\tau_u$, and $\tau_v$ are step sizes of iterations, and the proximal mappings are given as $\Pi_P({\bf p}) = \frac{{\bf p}}{ \max \{ 1, \|{\bf p}\| / \lambda_s\}}$ and  $\Pi_Q({\bf q}) = \frac{{\bf q}}{ \max \{ 1, \|{\bf q}\| / \lambda_a\}}$.
For initialization, we set $u^0=\bar{d}^{-1}$, ${\bf v}^0=0$, ${\bf p}^0=0$, ${\bf q}^0=0$, $\hat{u}^0=\bar{d}^{-1}$, $\hat{v}^0=0$.
This optimization efficiently runs in parallel and can run in real time on modern GPUs.

\setlength{\fboxsep}{0pt}
\begin{figure}[t]
\centering
\subfloat[Color image]{\includegraphics[width=40mm]{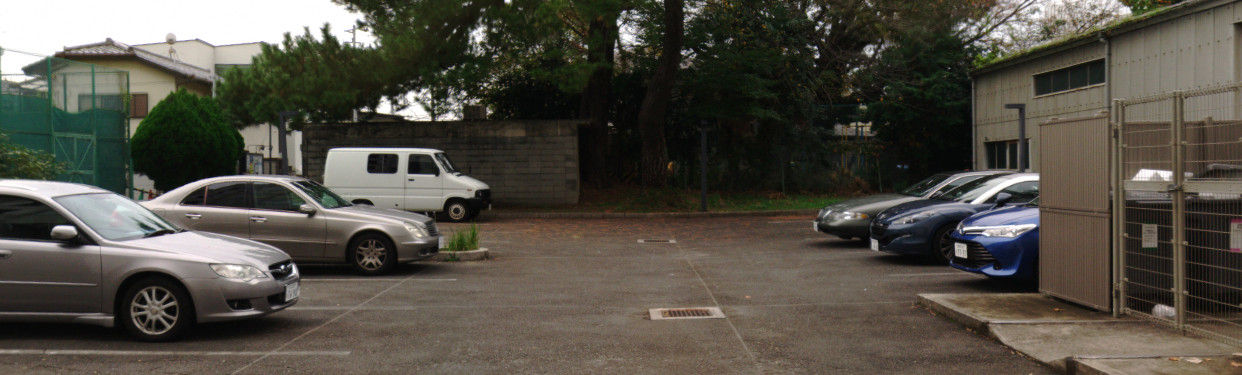}
}
\hfil
\subfloat[Semantic label]{\includegraphics[width=40mm]{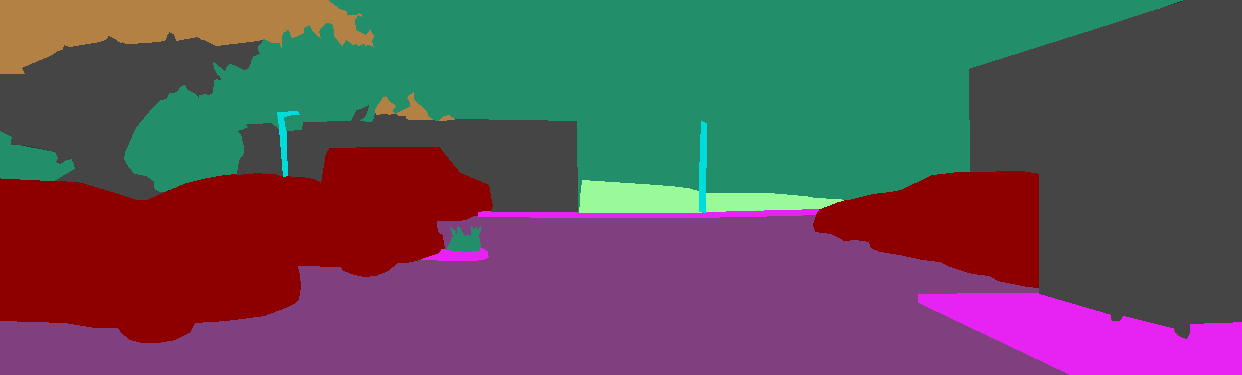}
}
\vspace{-8pt}
\subfloat[Dense depth map]{\fbox{\includegraphics[width=40mm]{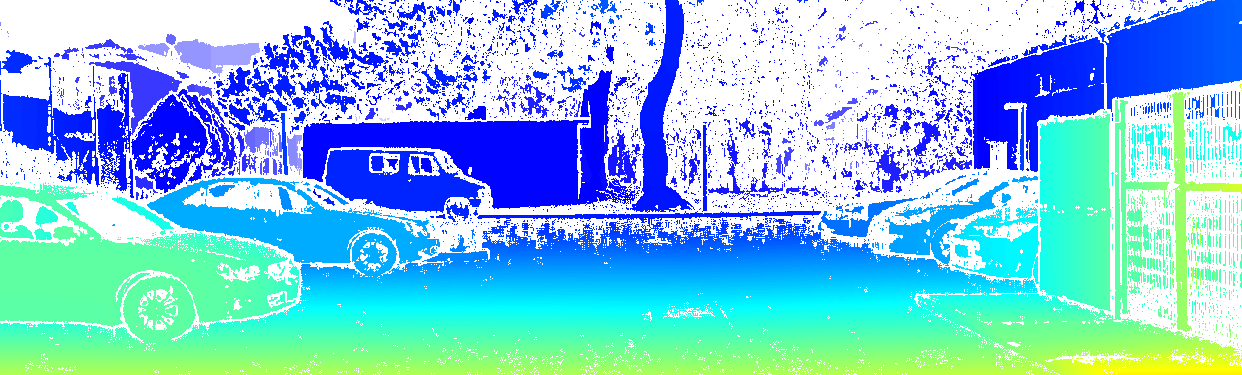}}
\label{fig_kitti_dense}
}
\hfil
\subfloat[Sparse depth map]{\fbox{\includegraphics[width=40mm]{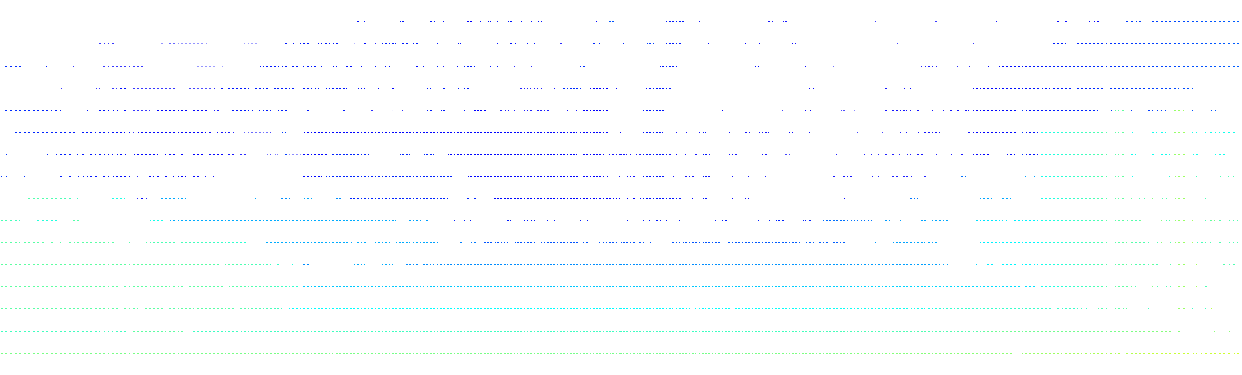}}
\label{fig_kitti_velo}
}
\caption{
A frame of Komaba. In depth maps, white indicates no value in data.
}
\label{fig_komaba}
\end{figure}

\begin{figure}[t]
\centering
\subfloat[Color image]{\includegraphics[width=27mm]{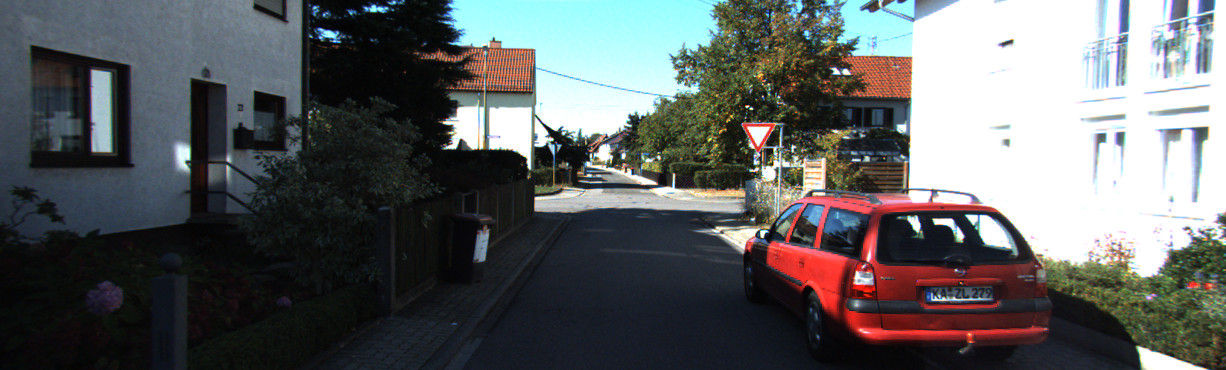}
}
\hfil
\subfloat[Semantic label]{\includegraphics[width=27mm]{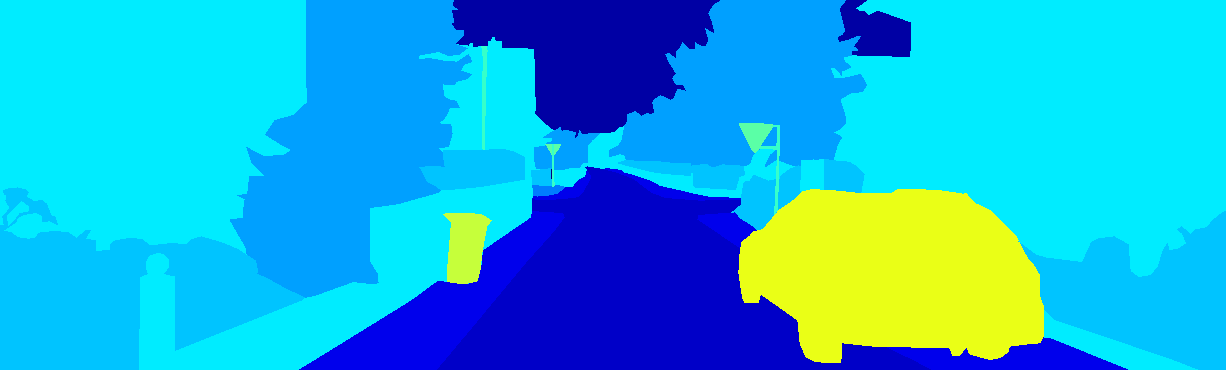}
}
\hfil
\subfloat[Dense depth map]{\fbox{\includegraphics[width=27mm]{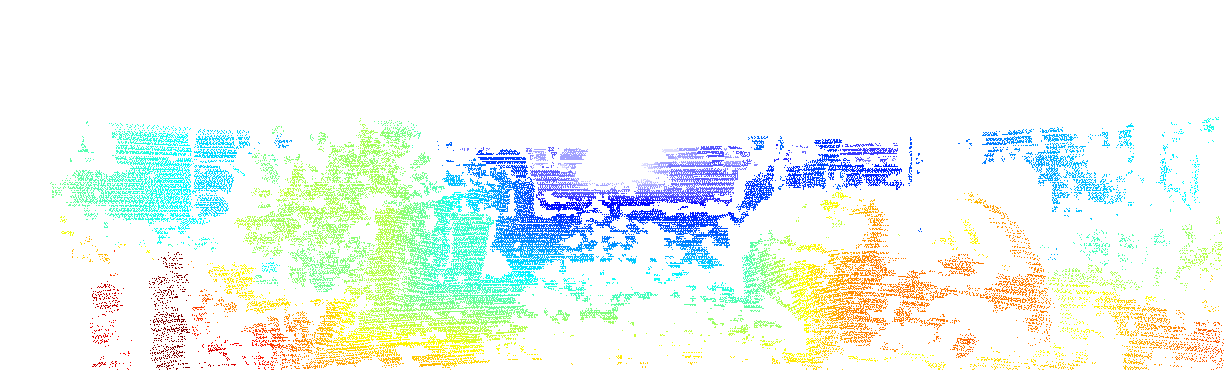}}
}
\vspace{-8pt}
\subfloat[Sparse depth map]{\fbox{\includegraphics[width=27mm]{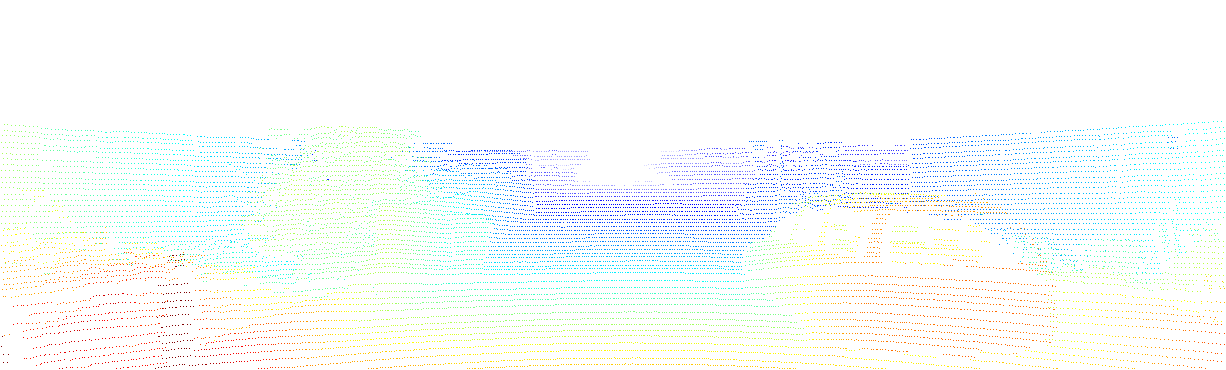}}
}
\hfil
\subfloat[Filtered sparse depth map]{\quad \fbox{\includegraphics[width=27mm]{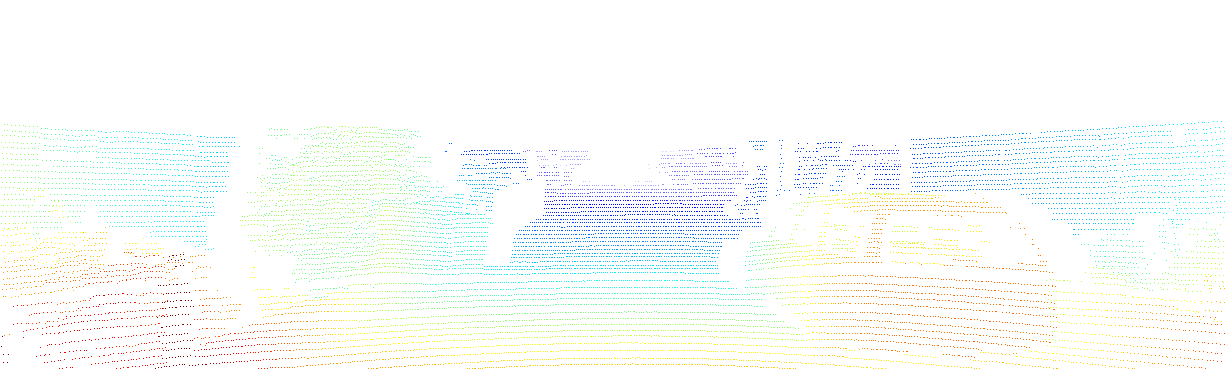}} \quad
}

\caption{
A frame of KITTI. In depth maps, white indicates no value in data.
}
\label{fig_kitti}
\end{figure}

\section{Evaluation}
\label{evaluation}
In this section, we evaluate our depth completion method on multiple aspects: depth map accuracy, processing time, point cloud quality, and parameter sensitivity.

\subsection{Datasets}
\label{sec_data}
We evaluated our method on two datasets, Komaba (Fig. \ref{fig_komaba}) and KITTI (Fig \ref{fig_kitti}).
For Komaba, we used 5 frames (Frame 8, 12, 18, 29, 35) following the evaluation of \cite{hirata2019real}.
For KITTI, we used 97 frames which are with semantic labeling by \cite{xu2016multimodal}.
Both datasets have colored images, sparse depth maps, dense depth maps, and manual semantic labels.

Because of the large disparity between the LIDAR and the camera, depth maps in KITTI contain occluded background noises.
Since our method does not handle such noises, we pre-processed the sparse depth maps to remove them (Fig. \ref{fig_kitti} (e)).
We used the filtered depth maps as input in the experiments on KITTI.

More details about the datasets and our occluded background noise removal approach are described in Appendix \ref{app_data}.

\subsection{Experimental Conditions}
\label{sec_baselines}
\paragraph{Baseline methods}
For the nearest neighbor search, we compared our method with two baselines.
One is the nearest neighbor search on distance only (NNS).
The other is Joint Nearest Neighbor Search (JNNS), which is based on Joint Distance (JD) of color and space.
JD is based on the metric for the interpolation of Kopf et al. \cite{kopf2007joint}.
JD from ${\bf i}\in\Omega$ to ${\bf j} \in \Omega$ is defined with weights $\alpha$ and $\beta$ as 
\begin{equation}
{\rm JD}({\bf i}, {\bf j}) = \alpha \left\|{\bf i} - {\bf j}\right\|^2 + \beta  \left(I({\bf i} ) - I({\bf j}) \right)^2 .
\end{equation}

For depth completion, we compared our methods with unsupervised methods of Kopf et al. \cite{kopf2007joint} and Ferstl et al. \cite{ferstl2013image} on both datasets, unsupervised method of Ku et al.'s \cite{ku2018defense} on KITTI only, and semi-supervised method of Hirata et al. \cite{ferstl2013image} on Komaba only.
Additionally, we improved Ferstl et al.'s method \cite{ferstl2013image} by using our data term $\bar{C}(u)$ instead of their $C(u)$ (``IGNNS+ADT").

We experimented with three versions of our method in terms of the use of the ground labels. 
The first is without occlusion boundary filtering (``Ours  no label'').
The second is with occlusion boundary filtering using the generated ground labels (``Ours'').
Refer Section \ref{sec_boundary} for our label generation.
The third is with occlusion boundary filtering using the manual labels in the datasets (``Ours man-label'').
Since ``Ours man-label'' uses additional inputs, the results are reference only and not used in evaluations.

\paragraph{Hyper parameters}
\label{app_cond}
We used the following parameters for our method on both datasets.
$c=0.01$ for IGNNS, $t=2.0$ [m] (meters) for the boundary derivation, and $\lambda_s=0.2$, $\lambda_a=1.6$, $\lambda_d=0.2$, $w=\bar{d}$, $\tau_p=\tau_q=1.0/\sqrt{8}$, $\tau_u=\tau_v=1.0/\sqrt{12}$ for the energy minimization.
$N_{\rm ransac}=1000$ and $t_{\rm ransac}=0.2$ [{\rm m}] for the ground detection.
The number of iterations is 400 for Komaba and 200 for KITTI.
This is because it took more iterations to converge for Komaba since the initial depth maps are sparser.

We used the following parameters on baseline methods.
For JNN and Kopf et al's method \cite{kopf2007joint}, we used $\alpha=100$, $\beta=0.2$.
For ADT aided methods, based on our tuning, we used $a=5.0$, $b=0.5$ for \cite{ferstl2013image}, and $a=10.0$, $b=0.5$ for ``IGNNS+ADT'', and the same parameters as our method for the energy minimization.
For Hirata et al.'s method \cite{hirata2019real}, we used the manual semantic labels and motion stereo provided in the dataset and conducted parameter tuning to adjust to our sampling.
Finally, for Ku et al's method \cite{ku2018defense}, we used their publicly available implementation as is, which was tuned for the KITTI depth completion benchmark by the authors.

Additionally, we used the following parameters for the pre-processing on KITTI.
$r _{\rm occ}= 256.0 * d^{-1}$ and $t_{\rm occ}=2.0$ [m] for the occluded background filtering (see Appendix \ref{app_data}).

\subsection{Depth Map Accuracy}
\label{sec_results}
We evaluated the accuracy by Mean Absolute Error (MAE) between the resulting depth maps and the ground truths in the datasets.
We show the nearest neighbor search results in Tables \ref{tab_komaba_nn} and \ref{tab_kitti_nn} and the depth completion results in Tables \ref{tab_komaba_dc} and \ref{tab_kitti_dc}.

In Tables \ref{tab_komaba_nn} and \ref{tab_kitti_nn}, we see IGNNS outperformed NNS and JNNS on both datasets.
In Tables \ref{tab_komaba_dc} and \ref{tab_kitti_dc}, we see our depth completion outperformed the previous and baseline methods with no ground label on both datasets (see ``Ours no label'').
Moreover, occlusion boundary filtering by our generated ground labels improved the accuracy further (see ``Ours'').

In Tables \ref{tab_komaba_dc} and \ref{tab_kitti_dc}, we see ``Ours'' is competitive to ``Ours man-label''.
The MAE differences between them are less than 5 [mm] on both datasets.
Furthermore, ``Ours'' outperformed ``Ours man-label'' on KITTI.
This result shows that the generated ground labels were as effective as the manual labels for our purpose. 

\begin{table}[t]
\caption{Nearest neighbor search results on Komaba dataset}
\centering
\begin{tabular}{|c|c|c|c|c|c|c|}
\hline
&\multicolumn{6}{|c|}{MAE [mm]} \\
\cline{2-7} 
& F. 8 & F. 12 & F. 18 & F. 29 & F. 35 & Avg. \\
\hline
NNS & 265.1 & 545.2 & 167.3 & 160.6 & 234.1 & 274.5 \\
\hline
JNNS & \textbf{253.2} & 546.8 & 168.2 & 160.5 & 246.4 & 275.0  \\
\hline
IGNNS &  267.1 & \textbf{433.5}  & \textbf{164.5} & \textbf{115.7} & \textbf{226.8} & \textbf{241.5}  \\
\hline
\end{tabular}
\label{tab_komaba_nn}
\end{table}

\begin{table}[t]
\caption{Nearest neighbor search results on KITTI dataset}
\centering
\begin{tabular}{|c|c|}
\hline
& MAE [mm] \\
\hline
NNS & 530.3 \\
\hline
JNNS &  527.9 \\
\hline
IGNNS&  \textbf{526.1}  \\
\hline
\end{tabular}
\label{tab_kitti_nn}
\end{table}

\begin{table}[t]
\caption{Depth completion results on Komaba dataset}
\centering
\begin{tabular}{|c|c|c|c|c|c|c|}

\hline
&\multicolumn{6}{|c|}{MAE[mm]} \\
\cline{2-7} 
& F. 8 & F. 12 & F. 18 & F. 29 & F. 35 & Avg. \\
\hline
\scriptsize
Kopf et al.
\cite{kopf2007joint} 
\normalsize
& 291.9 & 533.6 & 163.0 & 200.2 & 383.3 & 314.4 \\
\hline
\scriptsize
Ferstl et al.
 \cite{ferstl2013image}
 \normalsize
 & 603.9 & 1045.4  & 301.0 & 660.4 & 1094.1 & 740.9  \\
\hline
\scriptsize
Hirata et al.
 \cite{hirata2019real} 
\normalsize
& 1036.8 & 1096.2 & 421.7 & 443.2 & 711.9 & 742.0 \\
\hline
\scriptsize
IGNNS+ADT 
\normalsize
& 251.7 & 397.5 & 159.2 & 131.0 & 246.5 & 237.2 \\
\hline
\scriptsize
Ours no label 
\normalsize
& 255.9 & 383.8 & 135.8 & \textbf{116.7} & \textbf{216.6} & 221.8 \\

\hline
\scriptsize
Ours 
\normalsize
& \textbf{227.7} & \textbf{321.4} & \textbf{134.3} & 127.0 & 221.6 & \textbf{206.4} \\
\hline
\hline
\scriptsize
Ours man-label 
\normalsize
&228.7 & 321.3 & 134.0 & 116.9 & 213.7 & 202.9 \\
\hline
\end{tabular}
\label{tab_komaba_dc}
\end{table}

\begin{table}[t]
\caption{Depth completion results on KITTI dataset}
\centering
\begin{tabular}{|c|c|}
\hline
& MAE [mm] \\   
\hline
Kopf  et al. \cite{kopf2007joint} & 744.1 \\
\hline
Ferstl et al. \cite{ferstl2013image} & 928.7 \\
\hline
Ku et al. \cite{ku2018defense}  & 614.2 \\
\hline
IGNNS+ADT& 525.3 \\
\hline
Ours no label& 506.5 \\
\hline
Ours  & \textbf{500.2} \\
\hline \hline
Ours man-label & 504.1\\
\hline
\end{tabular}
\label{tab_kitti_dc}
\end{table}

\subsection{Processing time}
We implemented our method on an Ubuntu 18.04 LTS laptop computer running Intel(R) Core(TM) i9-8950HK @ 2.90GHz $\times$ 6 cores and GeForce RTX 2080 GPU. 
On average, the processing time per frame was 0.163 (IGNNS: 0.025, the ground label generation: 0.032, the energy minimization 0.107) second on Komaba and 0.120 (IGNNS: 0.023, the ground label generation: 0.070, the energy minimization 0.027) second on KITTI datasets.
The results indicate that it can process 6 - 8 frames per second on a single thread and can be applied to real-time applications with proper parallelization. 
Longer processing time for the energy minimization on Komaba is because the number of iterations is larger than on KITTI: 400 v.s. 200.
Ground detection took longer on KITTI than on Komaba because the input depth maps have more data.

\subsection{Point cloud quality}
\begin{figure*}[htbp]
\centering
\subfloat
{
\includegraphics[width=34mm]{figures/komaba_12/im12_img.jpg}
}
\hfil
\subfloat
{\includegraphics[width=34mm]{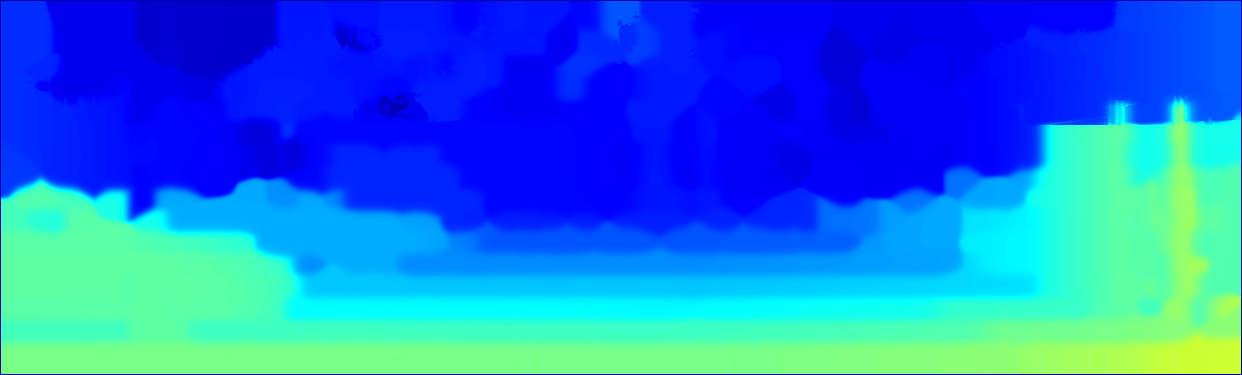}
}
\hfil
\subfloat
{\includegraphics[width=34mm]{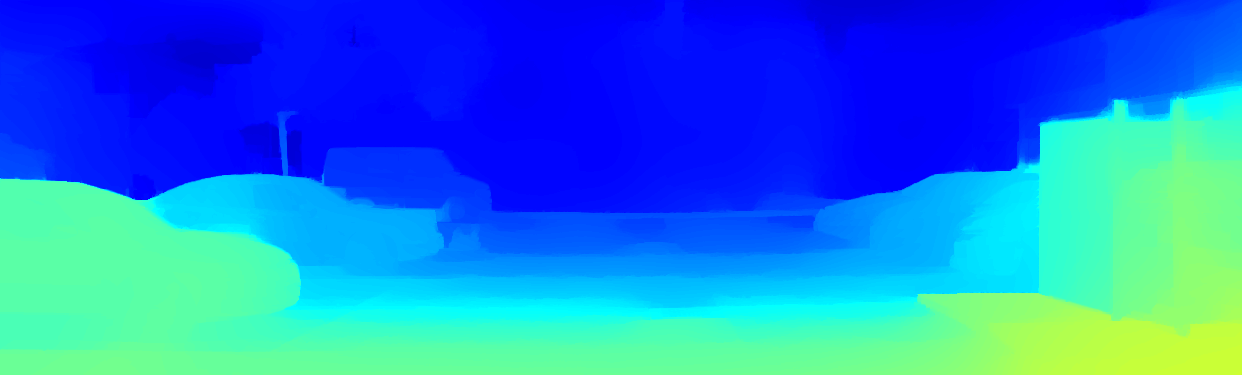}
}
\hfil
\subfloat
{\includegraphics[width=34mm]{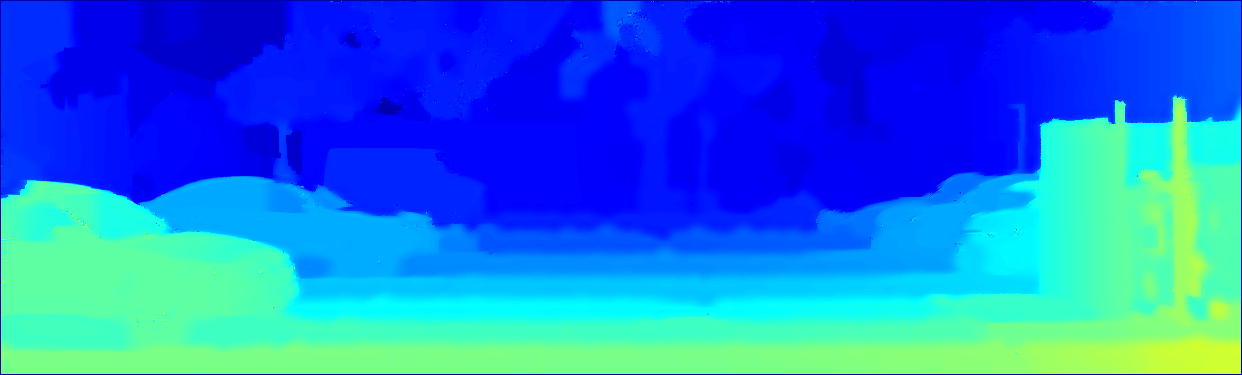}
}
\hfil
\subfloat
{\includegraphics[width=34mm]{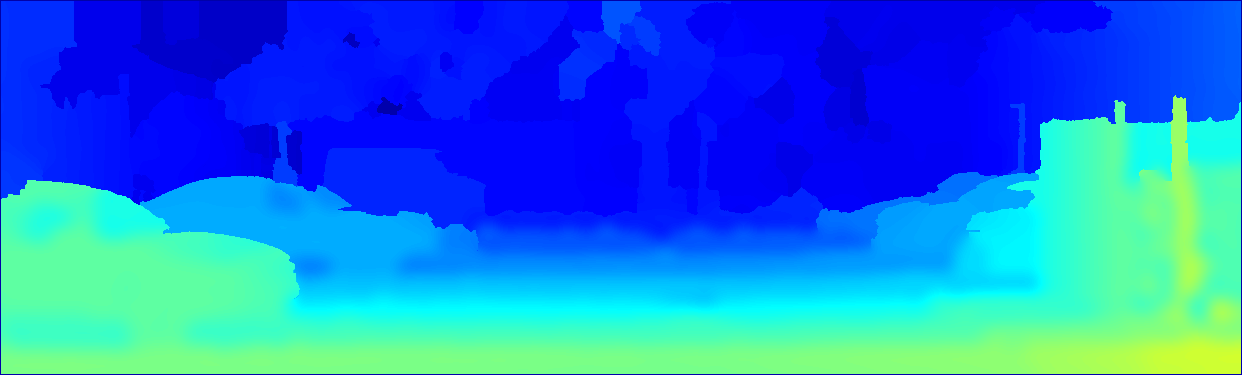}
}
\vspace{-8pt}
\subfloat
{
\includegraphics[width=34mm]{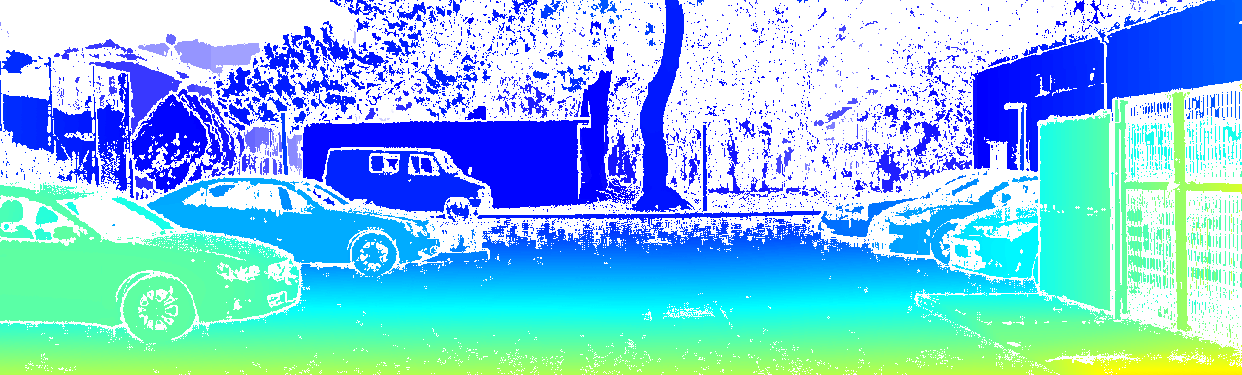}
}
\hfil
\subfloat
{\includegraphics[width=34mm]{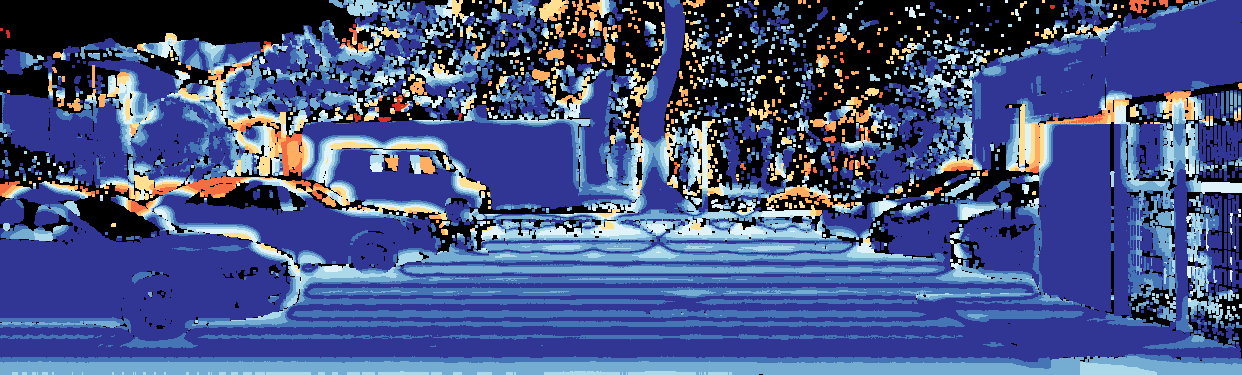}
}
\hfil
\subfloat
{\includegraphics[width=34mm]{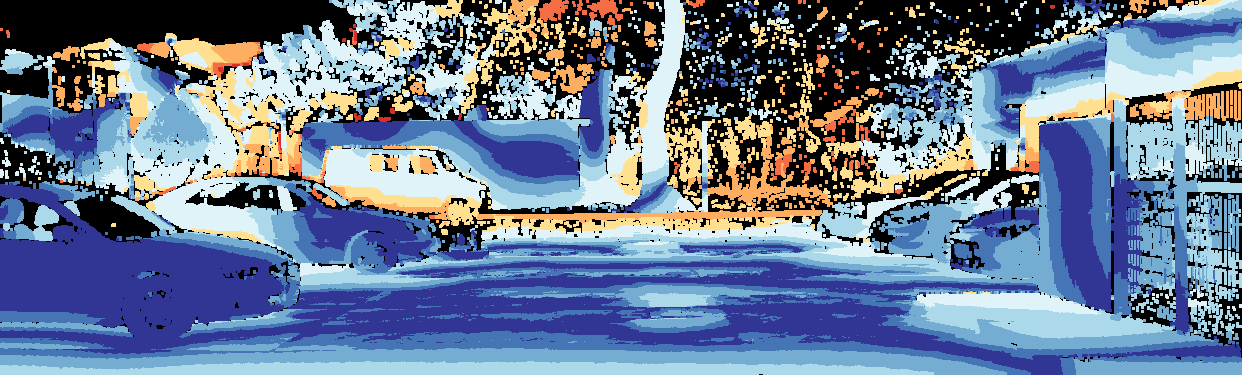}
}
\hfil
\subfloat
{\includegraphics[width=34mm]{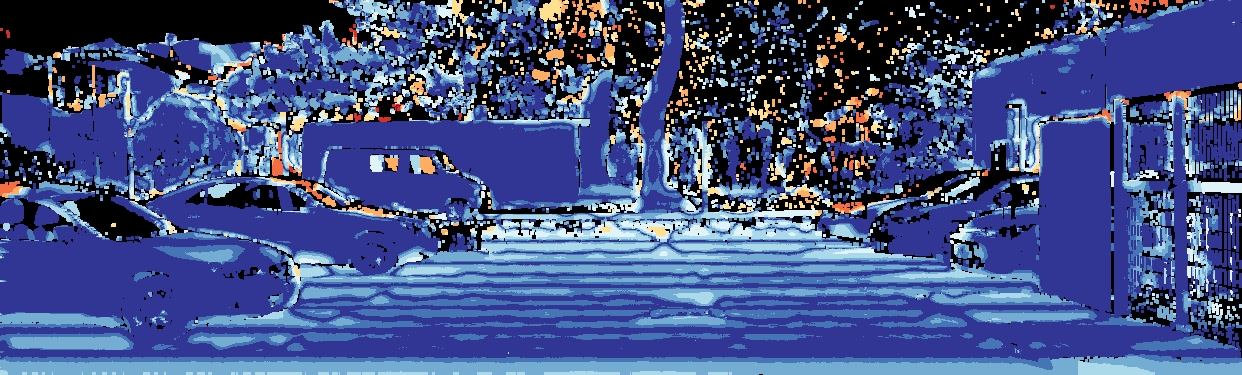}
}
\hfil
\subfloat
{\includegraphics[width=34mm]{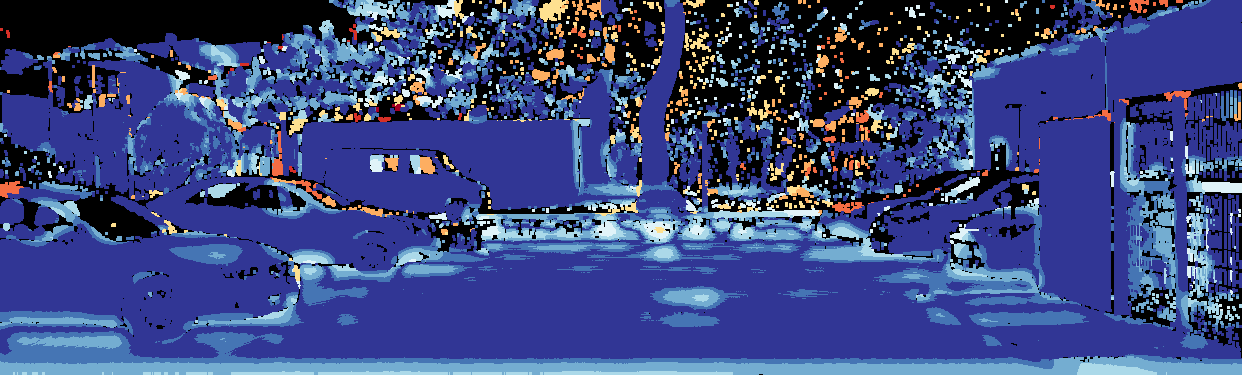}
}

\vspace{-8pt}
\setcounter{subfigure}{0}
\subfloat[Input and ground truth]{
\includegraphics[width=34mm]{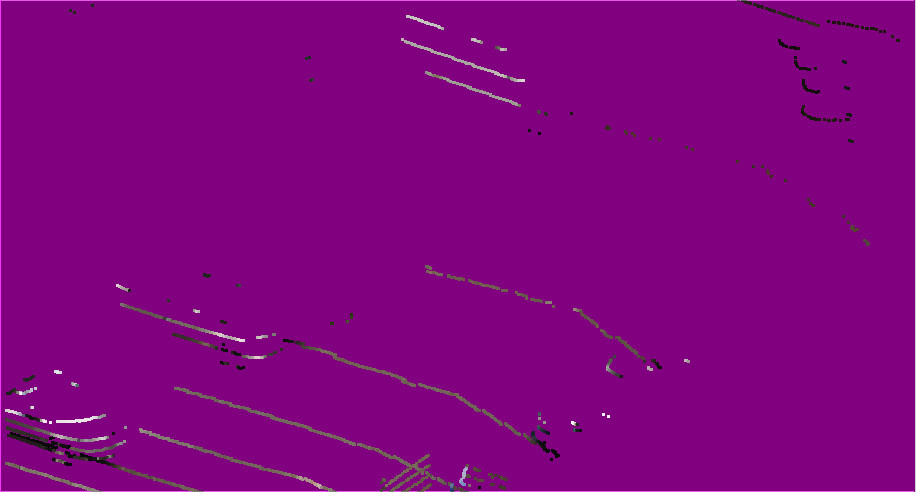}
}
\hfil
\subfloat[Result by \cite{kopf2007joint}]{\includegraphics[width=34mm]{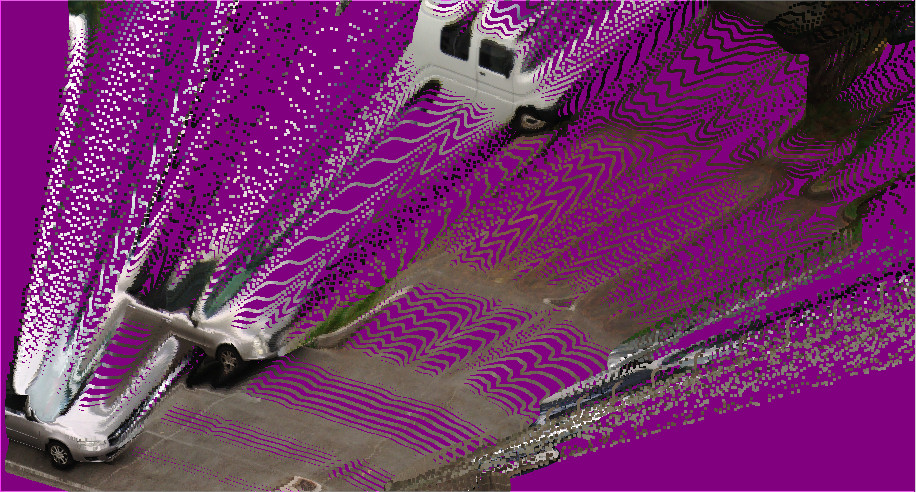}
}
\hfil
\subfloat[Result by \cite{hirata2019real}]{\includegraphics[width=34mm]{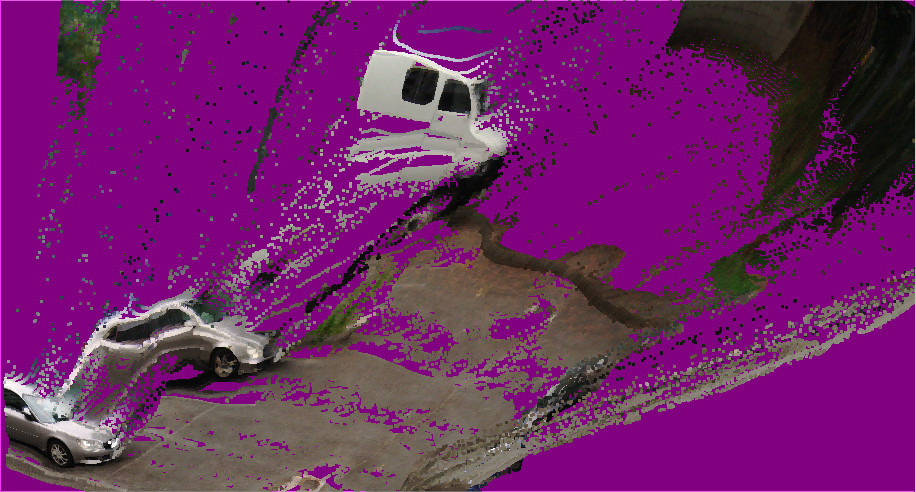}
}
\hfil
\subfloat[Result by IGNNS+ADT]{\includegraphics[width=34mm]{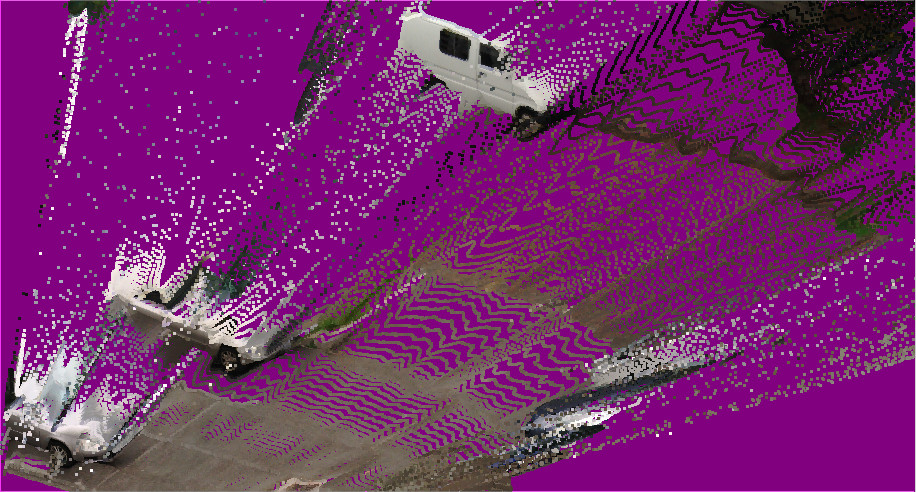}
}
\hfil
\subfloat[Result by Ours]{\includegraphics[width=34mm]{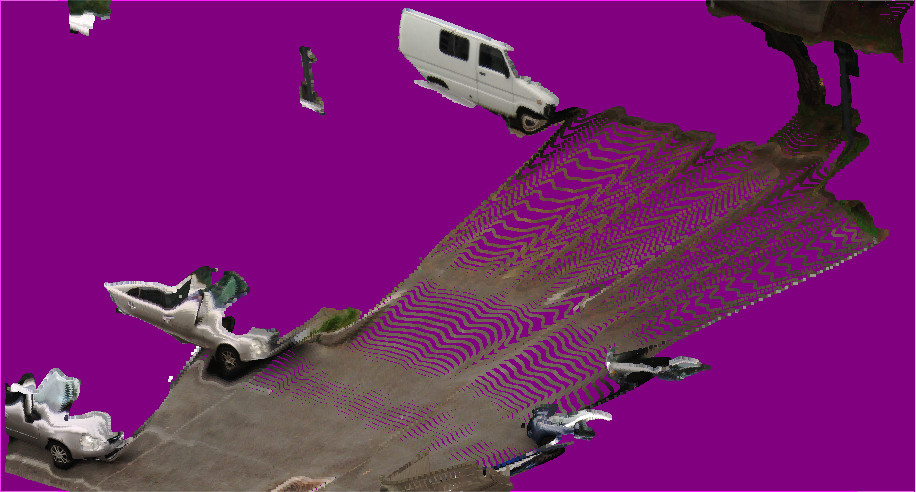}
}

\subfloat
{
\includegraphics[width=34mm]{figures/kitti_137/kitti_137_img.jpg}
}
\hfil
\subfloat
{\includegraphics[width=34mm]{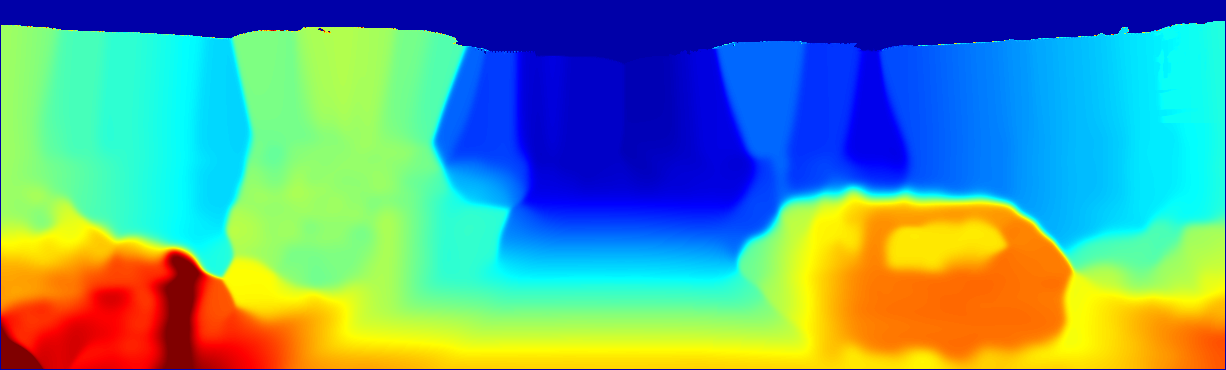}
}
\hfil
\subfloat
{\includegraphics[width=34mm]{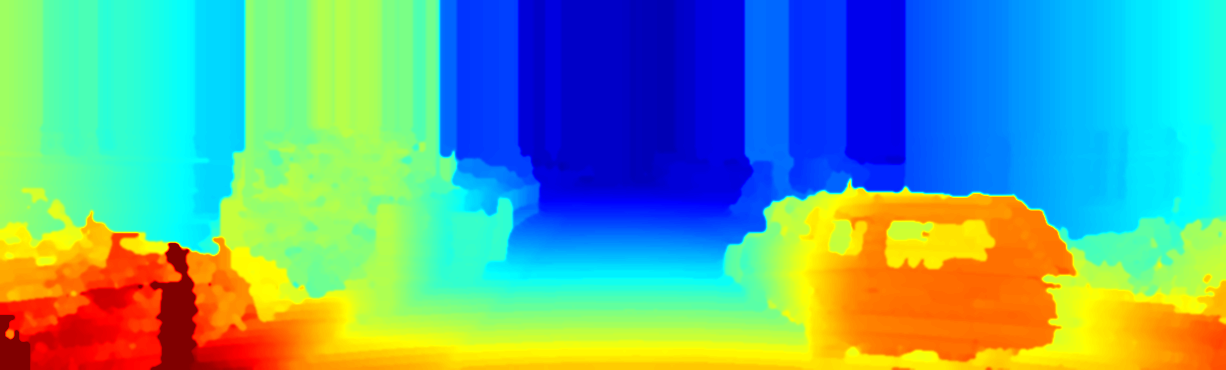}
}
\hfil
\subfloat
{\includegraphics[width=34mm]{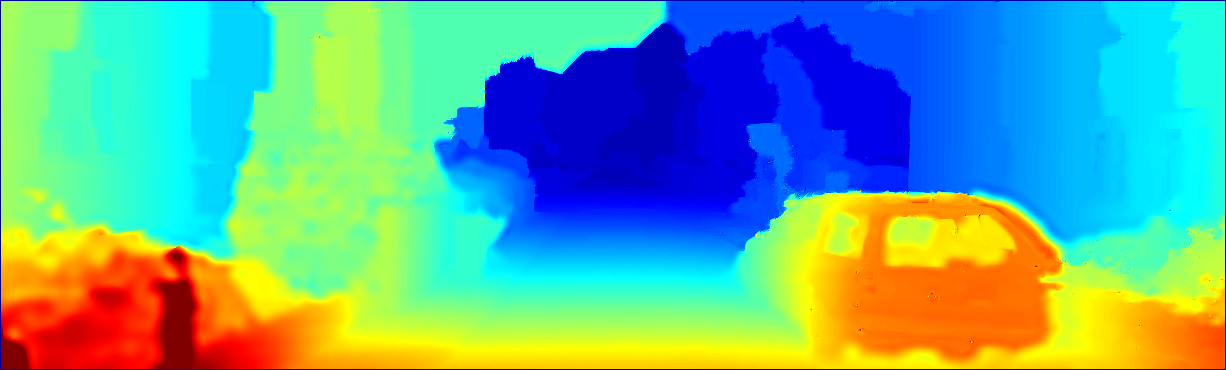}
}
\hfil
\subfloat
{\includegraphics[width=34mm]{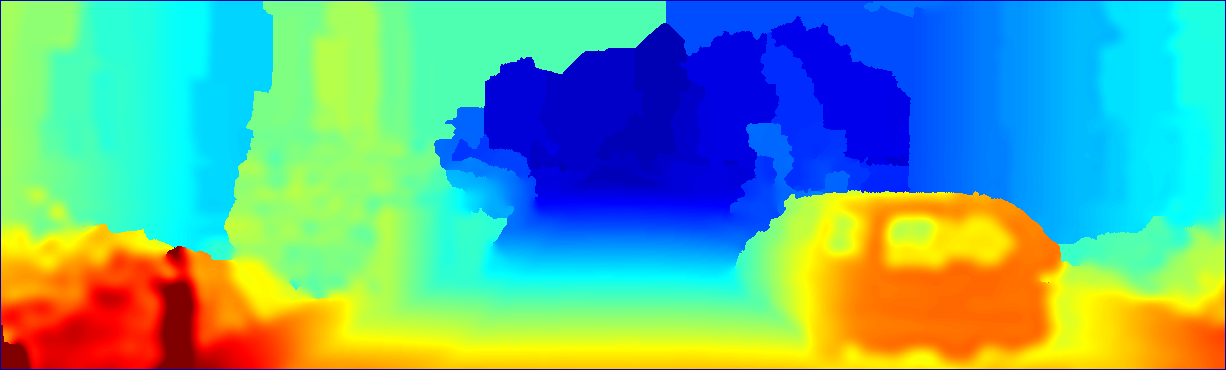}
}
\vspace{-8pt}

\subfloat
{
\includegraphics[width=34mm]{figures/kitti_137/kitti_137_gt_w2.png}
}
\hfil
\subfloat
{\includegraphics[width=34mm]{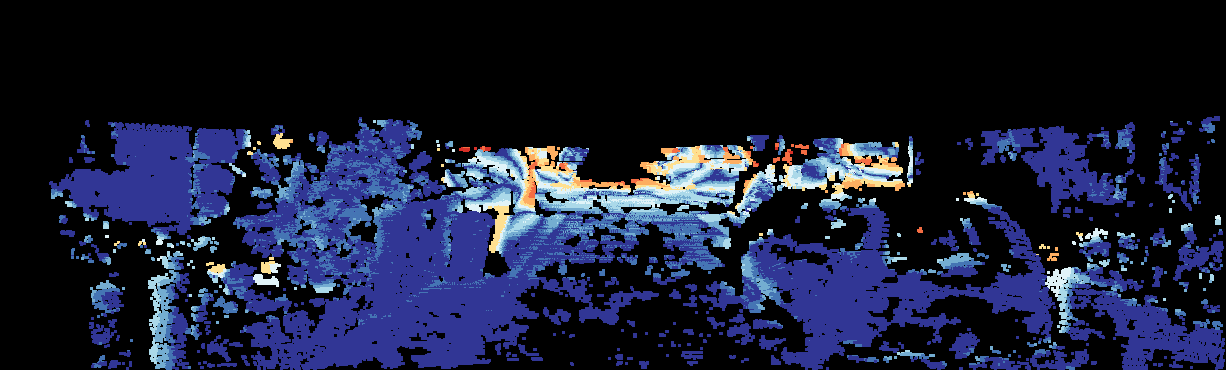}
}
\hfil
\subfloat
{\includegraphics[width=34mm]{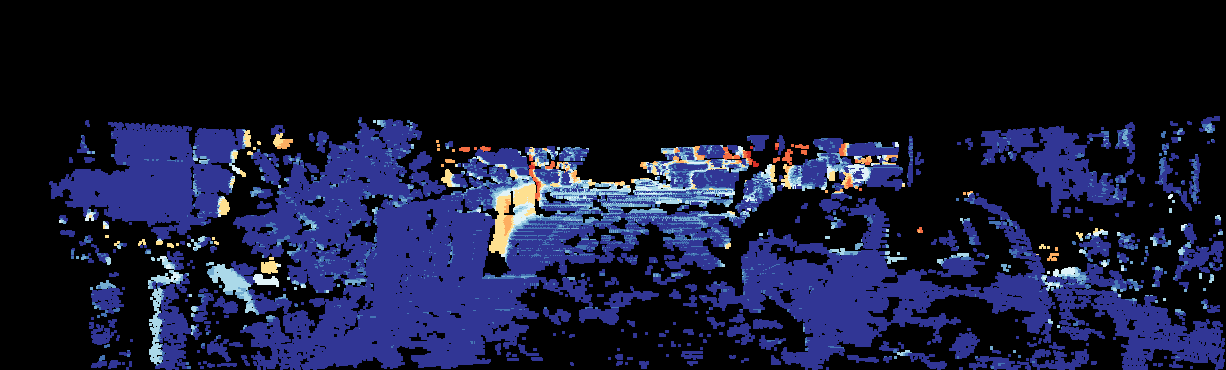}
}
\hfil
\subfloat
{\includegraphics[width=34mm]{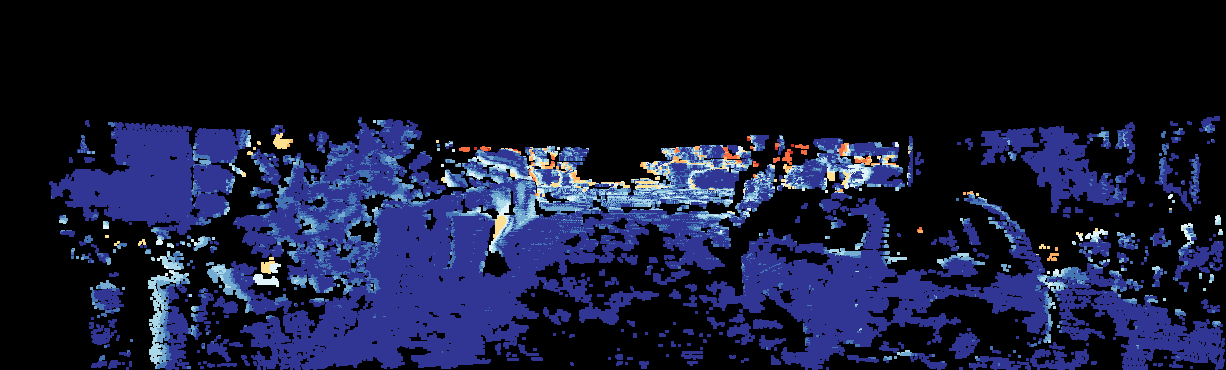}
}
\hfil
\subfloat
{\includegraphics[width=34mm]{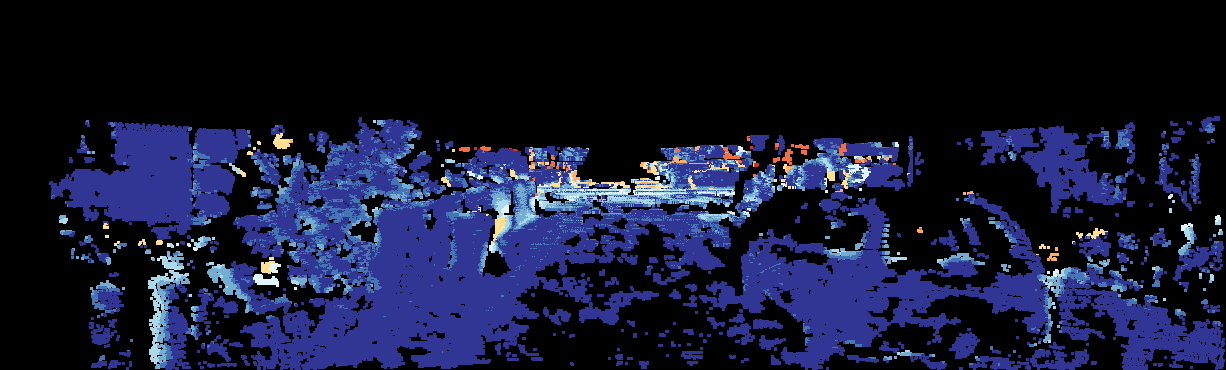}
}
\vspace{-8pt}
\subfloat
{
\includegraphics[width=34mm]{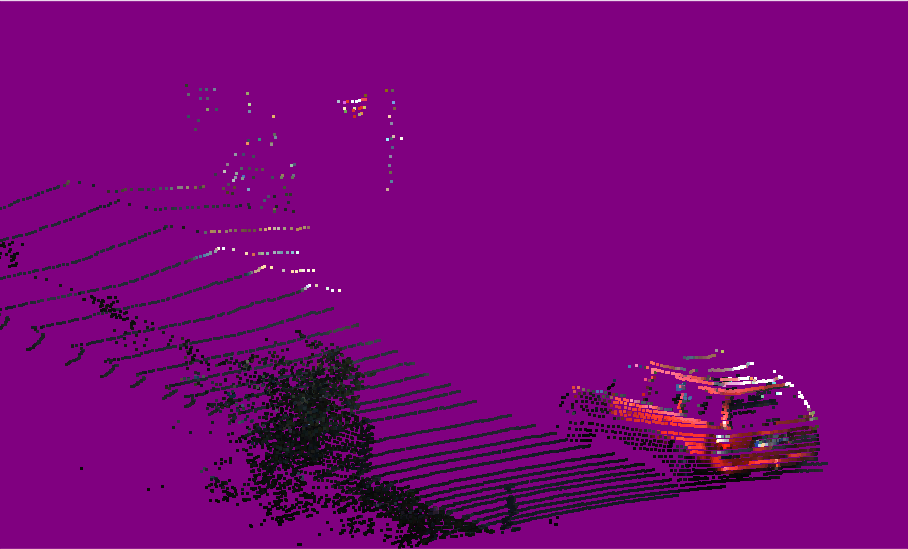}
}
\hfil
\subfloat
{\includegraphics[width=34mm]{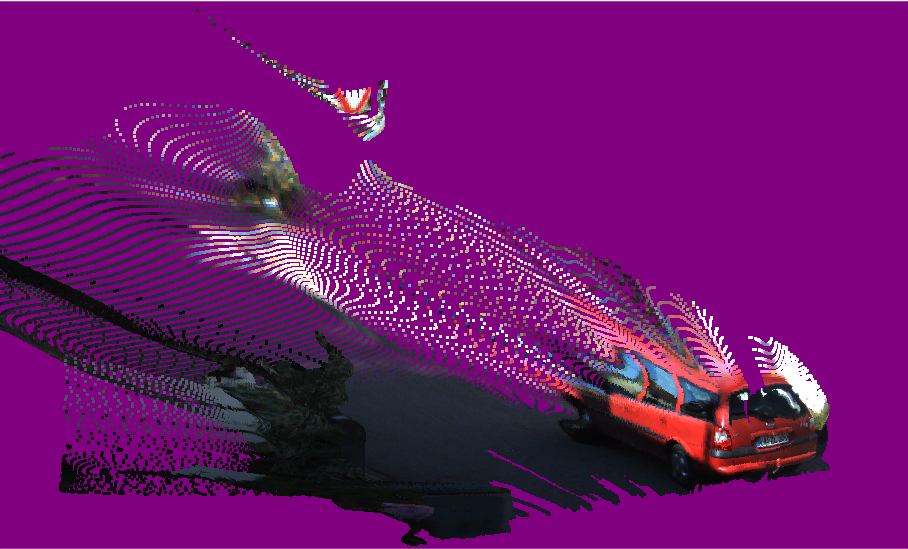}
}
\hfil
\subfloat
{\includegraphics[width=34mm]{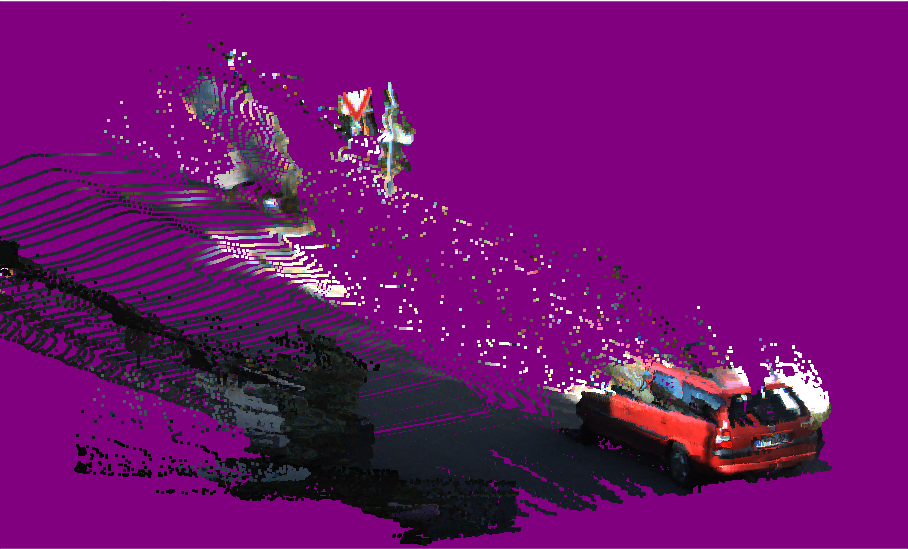}
}
\hfil
\subfloat
{\includegraphics[width=34mm]{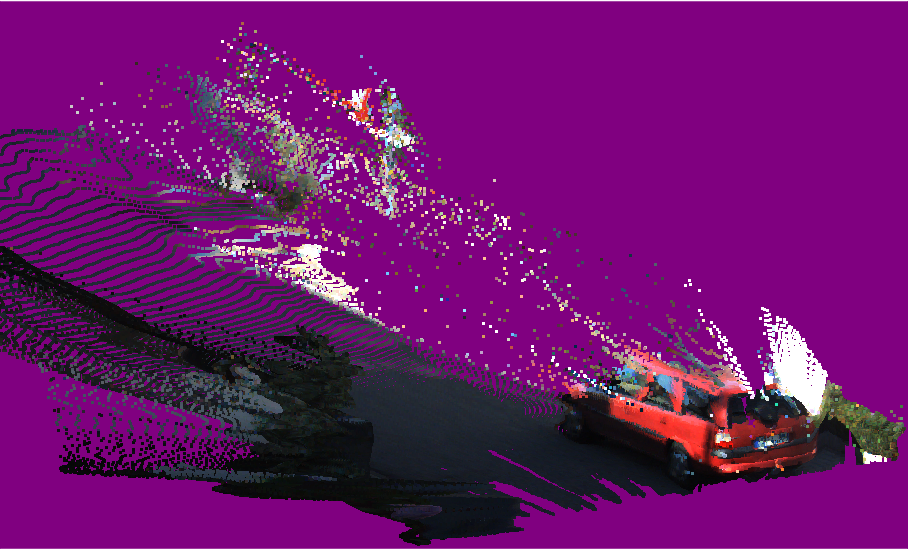}
}
\hfil
\subfloat
{\includegraphics[width=34mm]{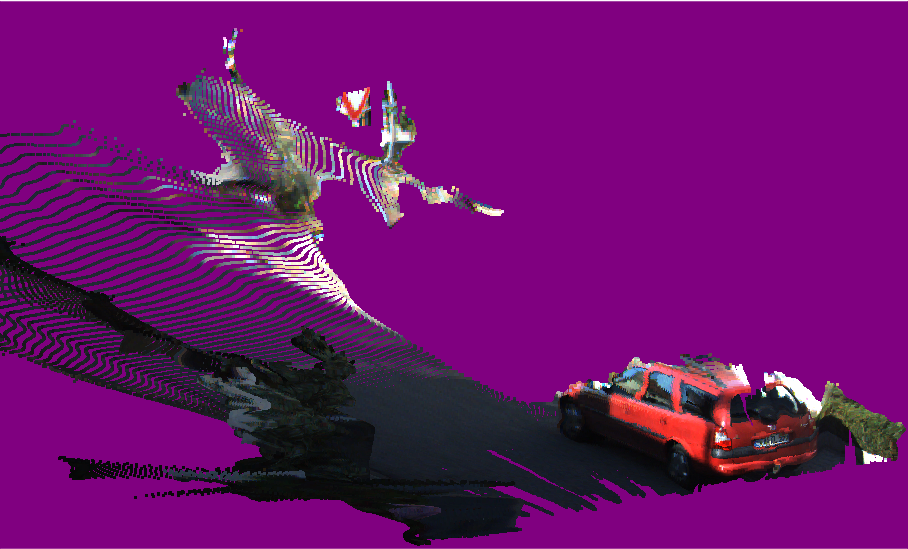}
}
\vfil
\subfloat
{
\includegraphics[width=34mm]{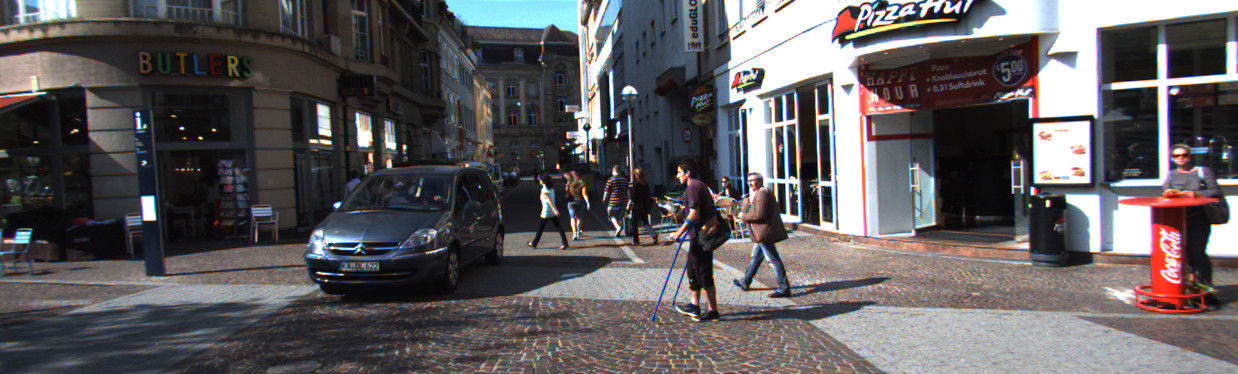}
}
\hfil
\subfloat
{\includegraphics[width=34mm]{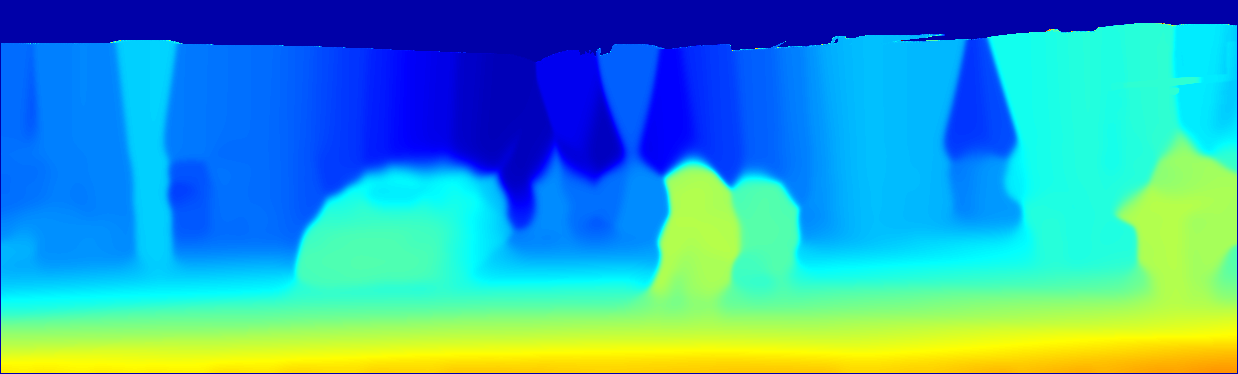}
}
\hfil
\subfloat
{\includegraphics[width=34mm]{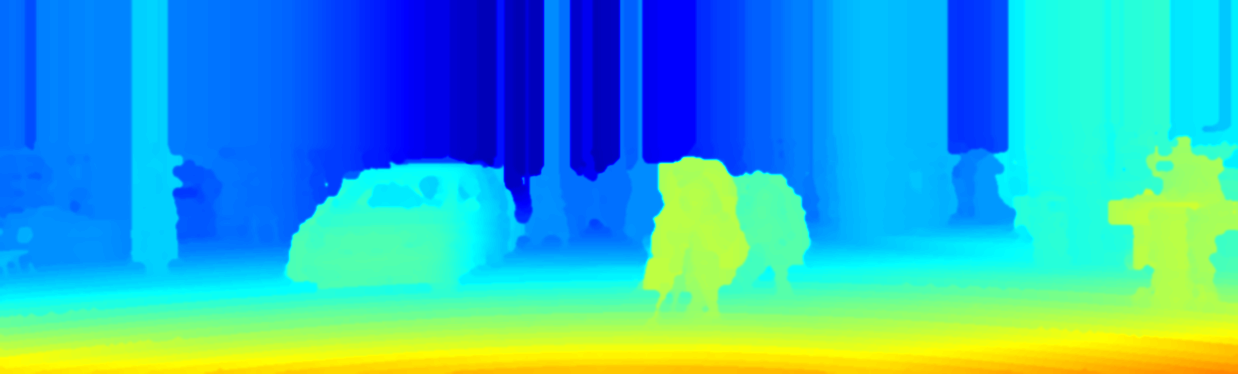}
}
\hfil
\subfloat
{\includegraphics[width=34mm]{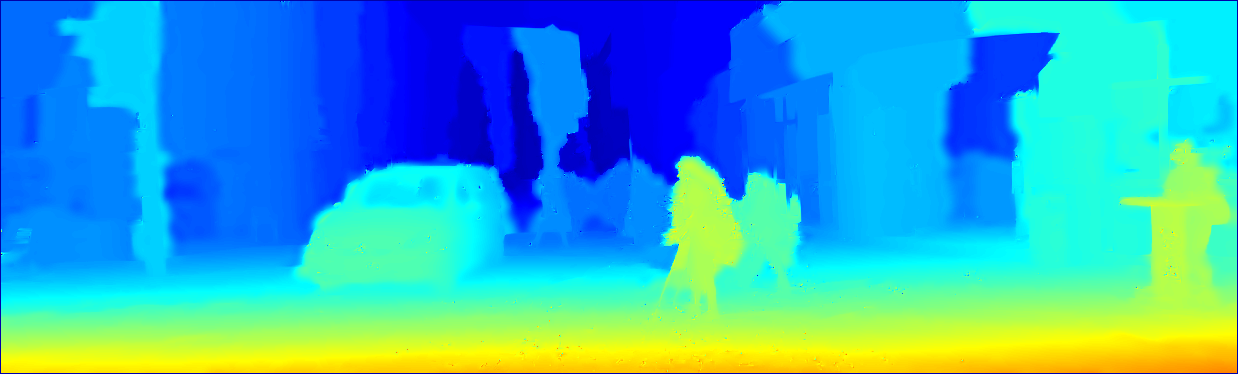}
}
\hfil
\subfloat
{\includegraphics[width=34mm]{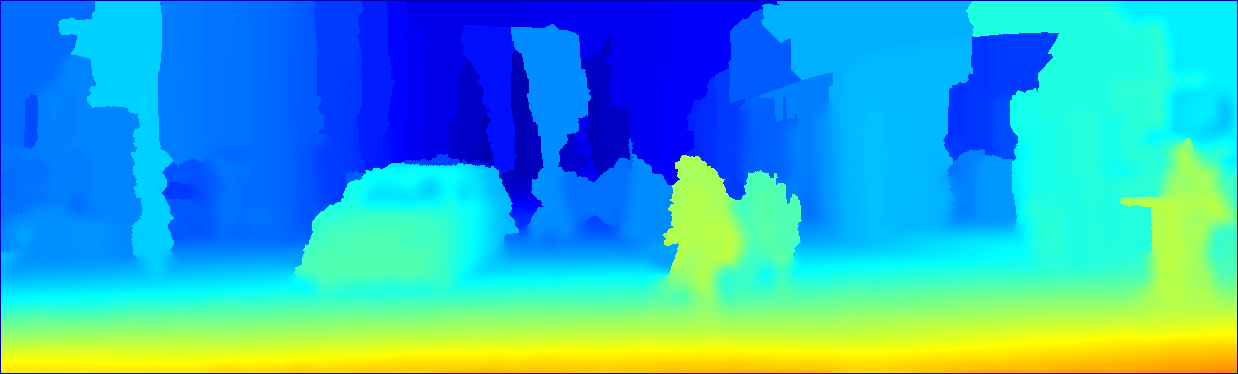}
}
\vspace{-8pt}
\subfloat
{
\includegraphics[width=34mm]{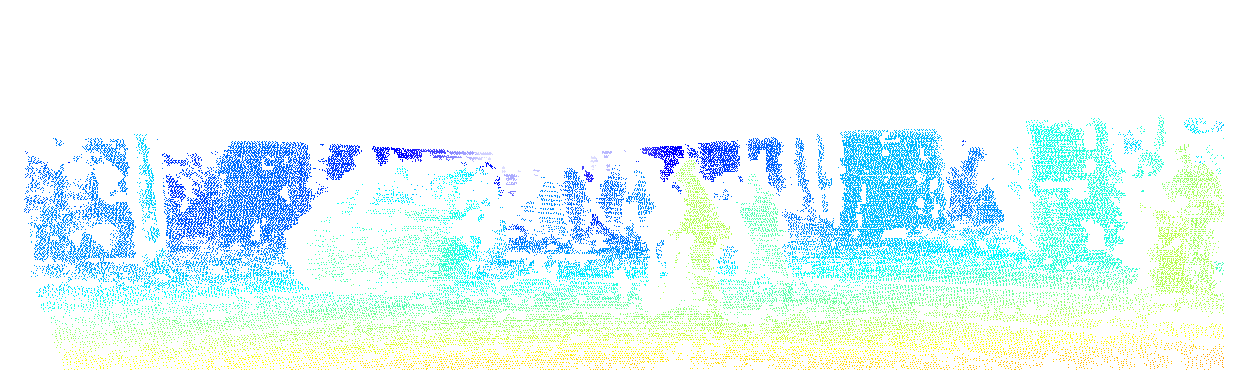}
}
\hfil
\subfloat
{\includegraphics[width=34mm]{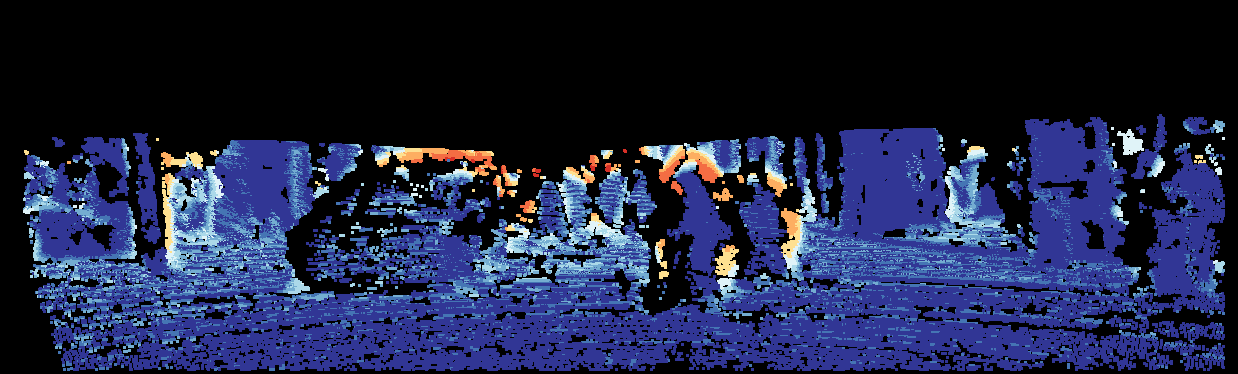}
}
\hfil
\subfloat
{\includegraphics[width=34mm]{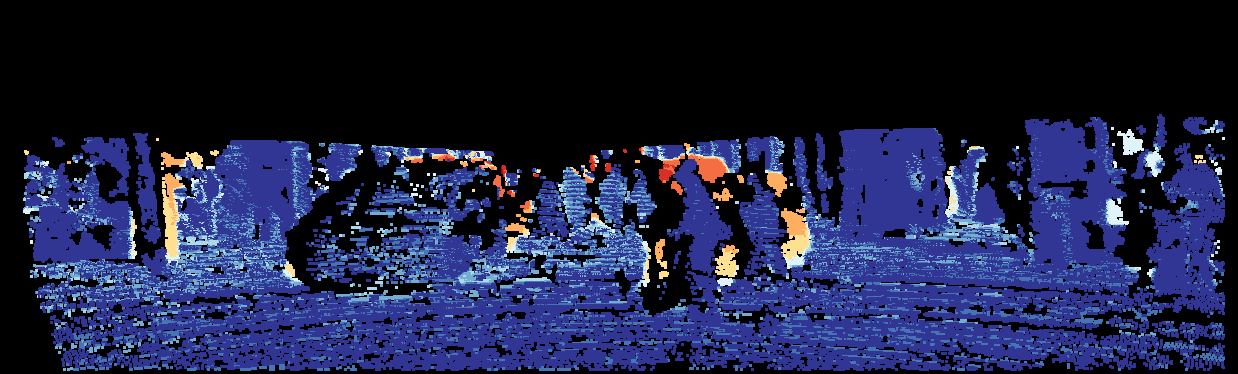}
}
\hfil
\subfloat
{\includegraphics[width=34mm]{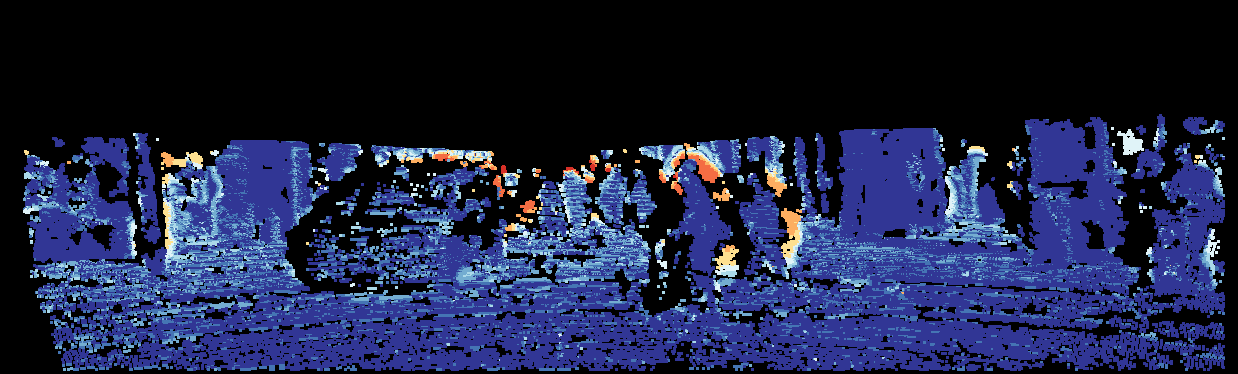}
}
\hfil
\subfloat
{\includegraphics[width=34mm]{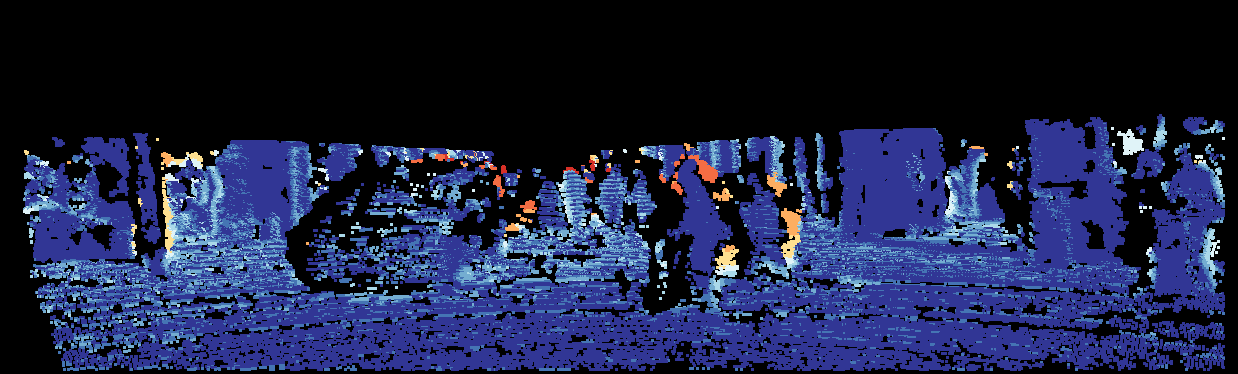}
}
\vspace{-8pt}
\setcounter{subfigure}{5}
\subfloat[Input and ground truth]{
\includegraphics[width=34mm]{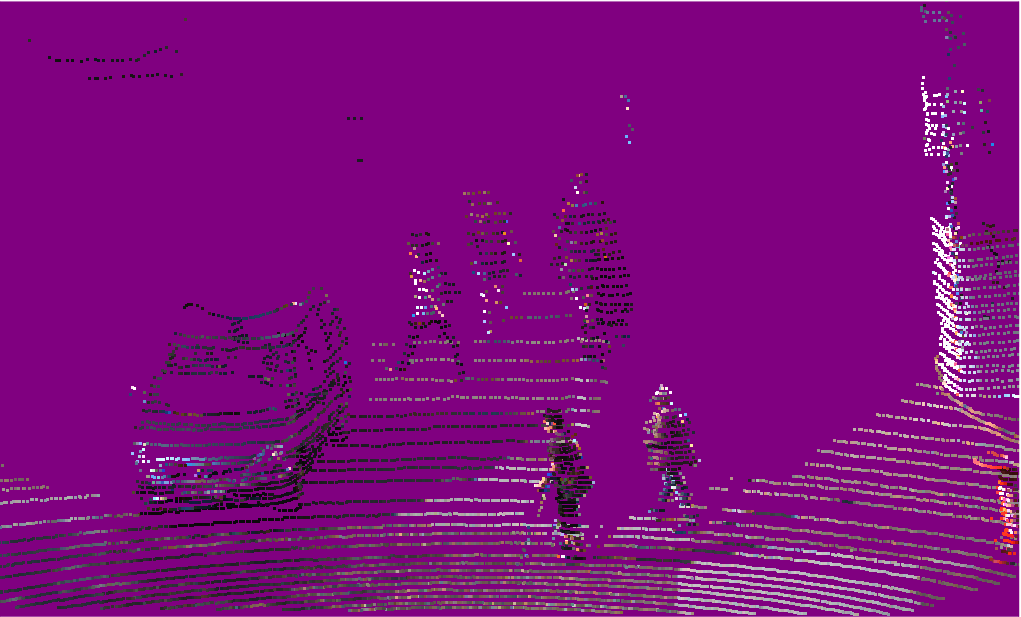}
}
\hfil
\subfloat[Result by \cite{kopf2007joint}]{\includegraphics[width=34mm]{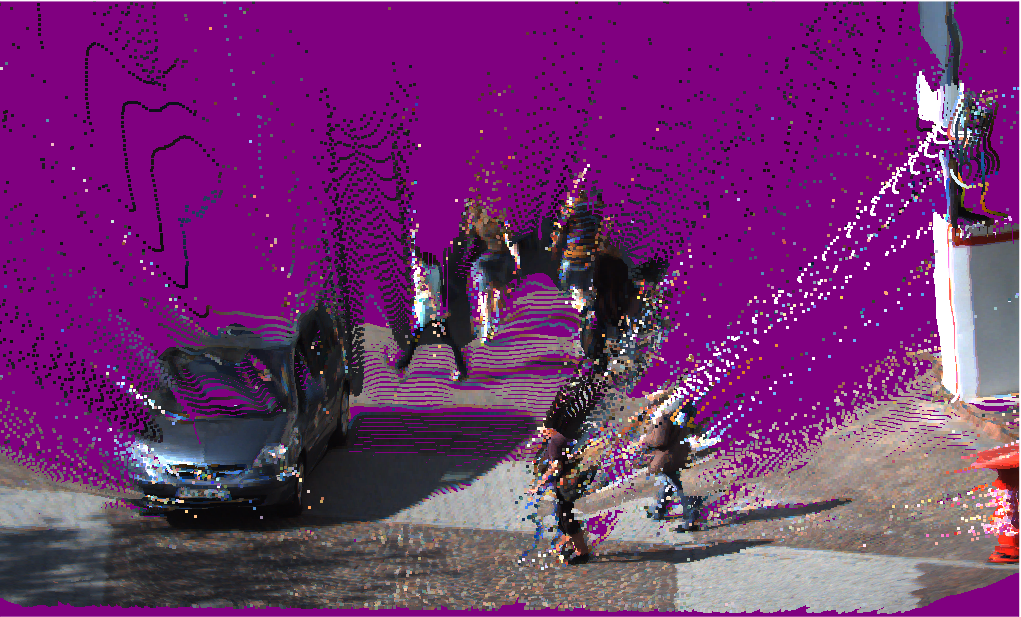}
}
\hfil
\subfloat[Result by \cite{ku2018defense}]{\includegraphics[width=34mm]{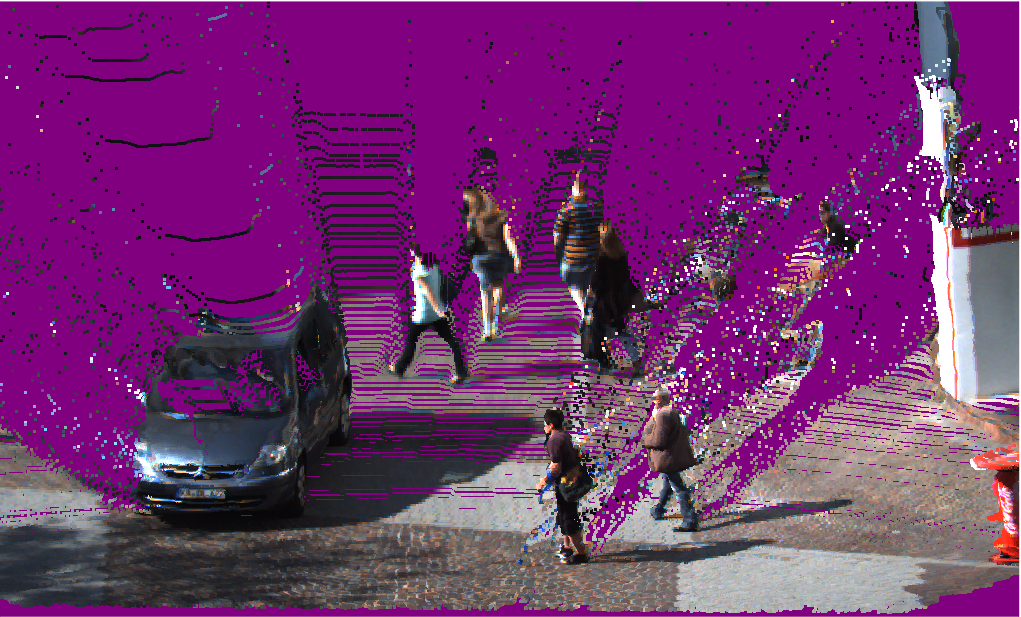}
}
\hfil
\subfloat[Result by IGNNS+ADT]{\includegraphics[width=34mm]{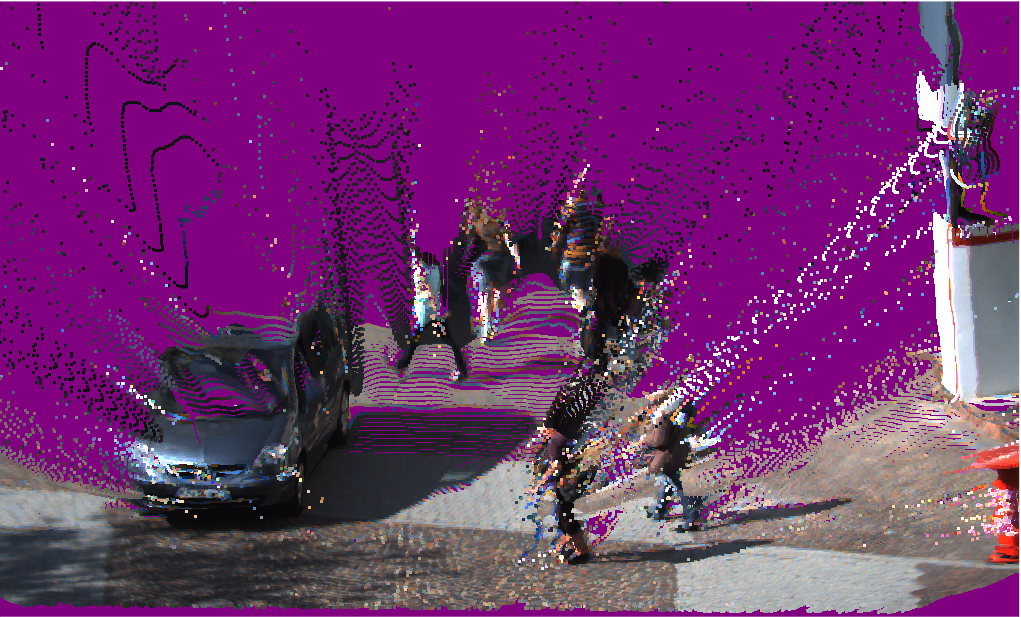}
}
\hfil
\subfloat[Result by Ours]{\includegraphics[width=34mm]{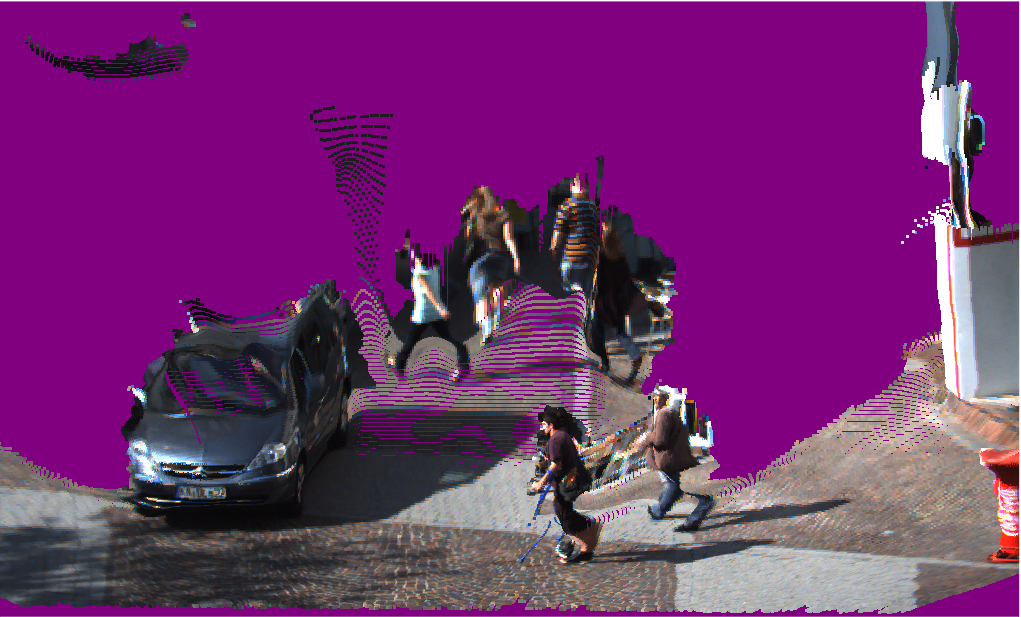}
}
\caption{
(a) - (e) Results on a selected frame of Komaba.
(f) - (j) Results on two selected frames of KITTI.
(a), (f) From top to bottom: the input image, the ground truth depth map, and the input LIDAR in point cloud.
(b) - (e), (g) - (j) From top to bottom: the depth completion result, the error map, and the depth completion result in point cloud.
By comparing the resulting point clouds, we see only our method preserves the discontinuity at occlusion boundaries. 
}
\label{fig_result_3}
\end{figure*}

We show selected results as depth maps and point clouds in Fig. \ref{fig_result_3}.
The advantage of our method is more obvious in point clouds than in depth maps.
In Fig. \ref{fig_result_3}, we can see that only our method preserves the discontinuity at occlusion boundaries. 
Other methods generate unnatural artifacts between foreground and background objects because of the smoothness imposed at the boundary during interpolation.
We can also see the effect in the error maps.
Results from other methods show gradual changes in the error around object boundaries, whereas our results show immediate changes.

\subsection{Parameter Sensitivity}
Our method introduces two hyper parameters: $c$ for IGNNS and $t$ for the boundary derivation.
We conducted parameter sensitivity evaluation on these. 
Here, we used ``Ours no label'' as a baseline to see the effects of the variables directly, and the other variables are fixed to values in Section \ref{sec_baselines}.

\begin{figure}[t]
\subfloat[Sesitivity to $c$]{\includegraphics[width=43mm]{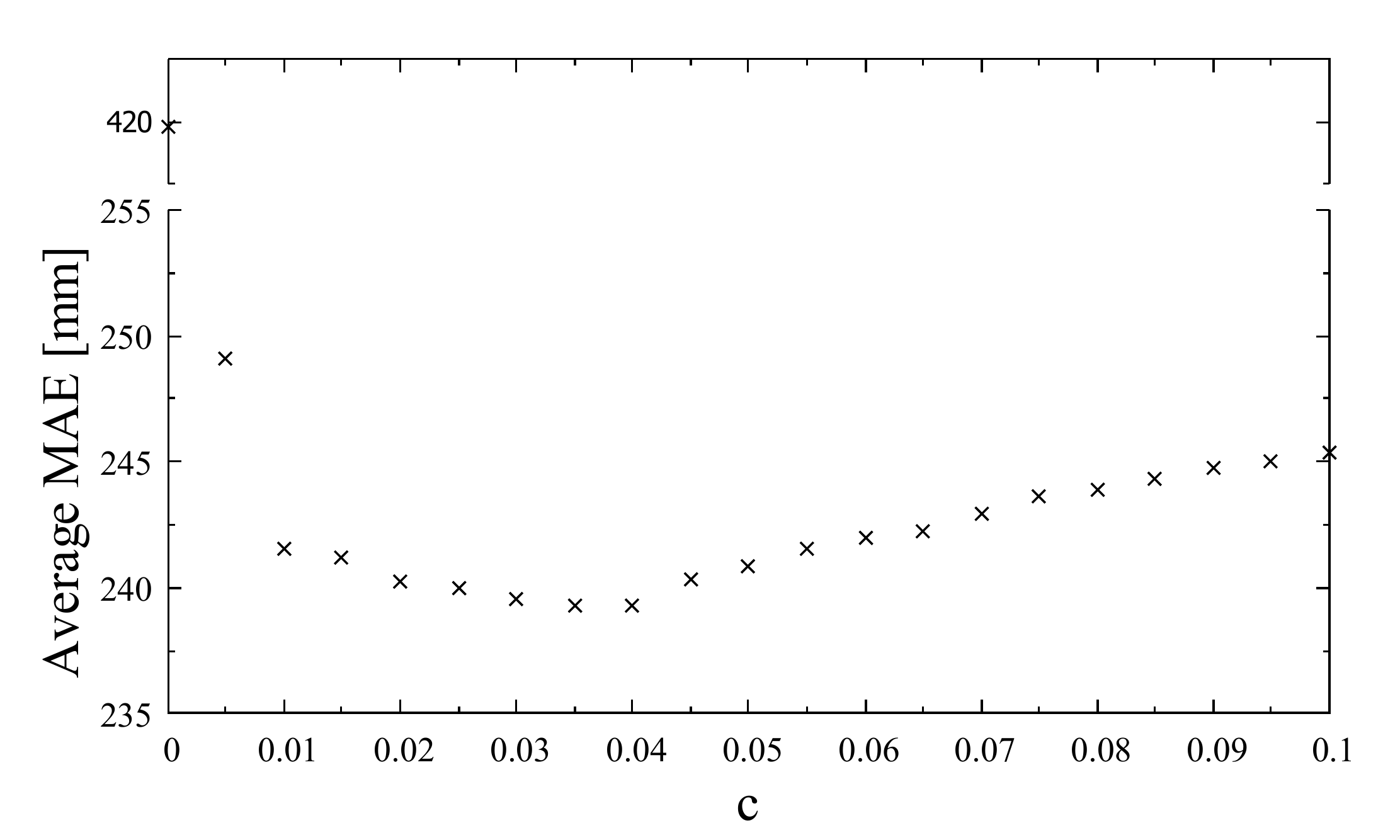}}
\hfil
\subfloat[Sensitivity to $t$]{\includegraphics[width=43mm]{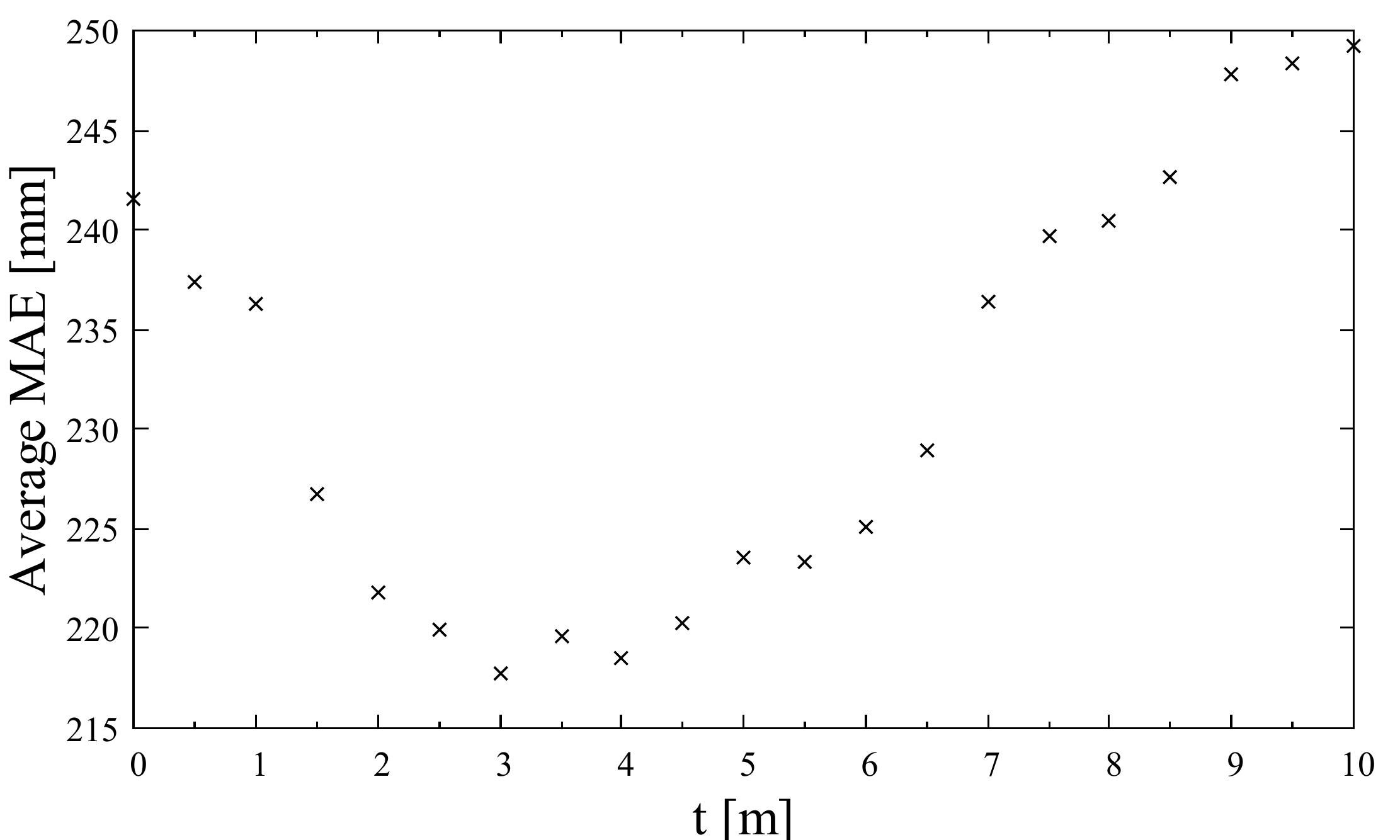}}
\caption{
Parameter sensitivity results on the Komaba dataset. MAE is calculated (a) on IGNNS results (b) after energy minimization.
}
\label{fig_sens}
\end{figure}

In Fig. \ref{fig_sens}, we plot the average MAE on Komaba dataset with various values of $c$ or $t$.
For $c$, MAE was the minimum ($239.3$) at $c=0.04$, and was less than or equal to the result of IGNNS in Table \ref{tab_komaba_nn} when $0.01 \leq c \leq 0.055$.
For $t$, MAE was the minimum ($217.7$) at $t=3.0$, and was less than or equal to the result of ``Ours no label'' in Table \ref{tab_komaba_dc} when $2.0 \leq t \leq 4.5$.

Although good parameters are data-dependent, our observation suggests the following for reasonable parameter choices.
Set $c$ to be on the scale of $1/100$ to $1/10$ against the square of the maximum color value, which in our case is $1.0$. 
Set $t$ to be on the scale of distance between the foreground and the background of the scene, e.g. several meters for the outdoor environments and several centimeters for the desktop environment.

\section{Benchmark Evaluation}
\label{sec_benchmark}
We applied our method to the public KITTI depth completion benchmark \cite{uhrig2017sparsity} to compare our method with various methods.
In the benchmark, there are the validation data for the tuning and the test data for the evaluation.
Both of them contain 1,000 pairs of an image and a sparse depth map.
The parameters used in the benchmark are the same as ``Ours" in Section \ref{sec_baselines} except for the following two parameters: $w=\bar{d}^{2.5}$ in the energy minimization and  $r_{\rm occ} = 96.0 * d^{-1}$ in the pre-processing.

In Table \ref{tab_test}, we show our results with all unsupervised \cite{ku2018defense} and DNN enhanced self-supervised \cite{ma2019self, wong2020unsupervised} methods listed in the benchmark website at the time of writing.
The results on the validation data are based on their published papers.
Among them, ours performed the best in terms of MAE.

\begin{table}[t]
\caption{Results on the benchmark}
\label{tab_test}
\centering
\begin{tabular}{|c|c|c|c|}
\hline
& \multicolumn{2}{|c|}{MAE [mm]} \\
\cline{2-3}
& validation & test \\
\hline
Ma  et al. \cite{ma2019self} &  358.9 &350.3  \\
\hline
Ku  et al. \cite{ku2018defense}  & 305.3 &302.6 \\
\hline
Wong et al.  \cite{wong2020unsupervised} & 305.1 &299.4  \\
\hline
Ours & \textbf{300.7}  &\textbf{298.7}  \\
\hline
\end{tabular}
\end{table}

\section{Conclusion}
We proposed B-ADT and its application to depth completion with IGNNS.
Our method is fully unsupervised, runs in real time, and generates a depth map discontinuous at occlusion boundaries and smooth elsewhere, which is suitable for point cloud representation.

We showed that our method outperformed both existing and baseline methods in terms of the accuracy of the depth maps and the visual quality of the point clouds.
Among the methods we evaluated, our method is the only one to preserve the discontinuity between foreground and background objects.

In this paper, we focused on B-ADT for depth completion.
However, B-ADT is a general idea that enables discontinuous but smooth optimization in variational energy minimization.
Variational methods have been applied to many tasks such as stereo matching, optical flow estimation, segmentation, and so on.
We believe it will be interesting to explore the possibility of B-ADT in many applications in the future.

\bibliographystyle{IEEEtran}
\bibliography{IEEEabrv, iros_2020_yao}

\appendices
\section{Convergence study}
\label{ap_conv}
\begin{figure}[t]
\subfloat[B-ADT only]{\includegraphics[width=43mm]{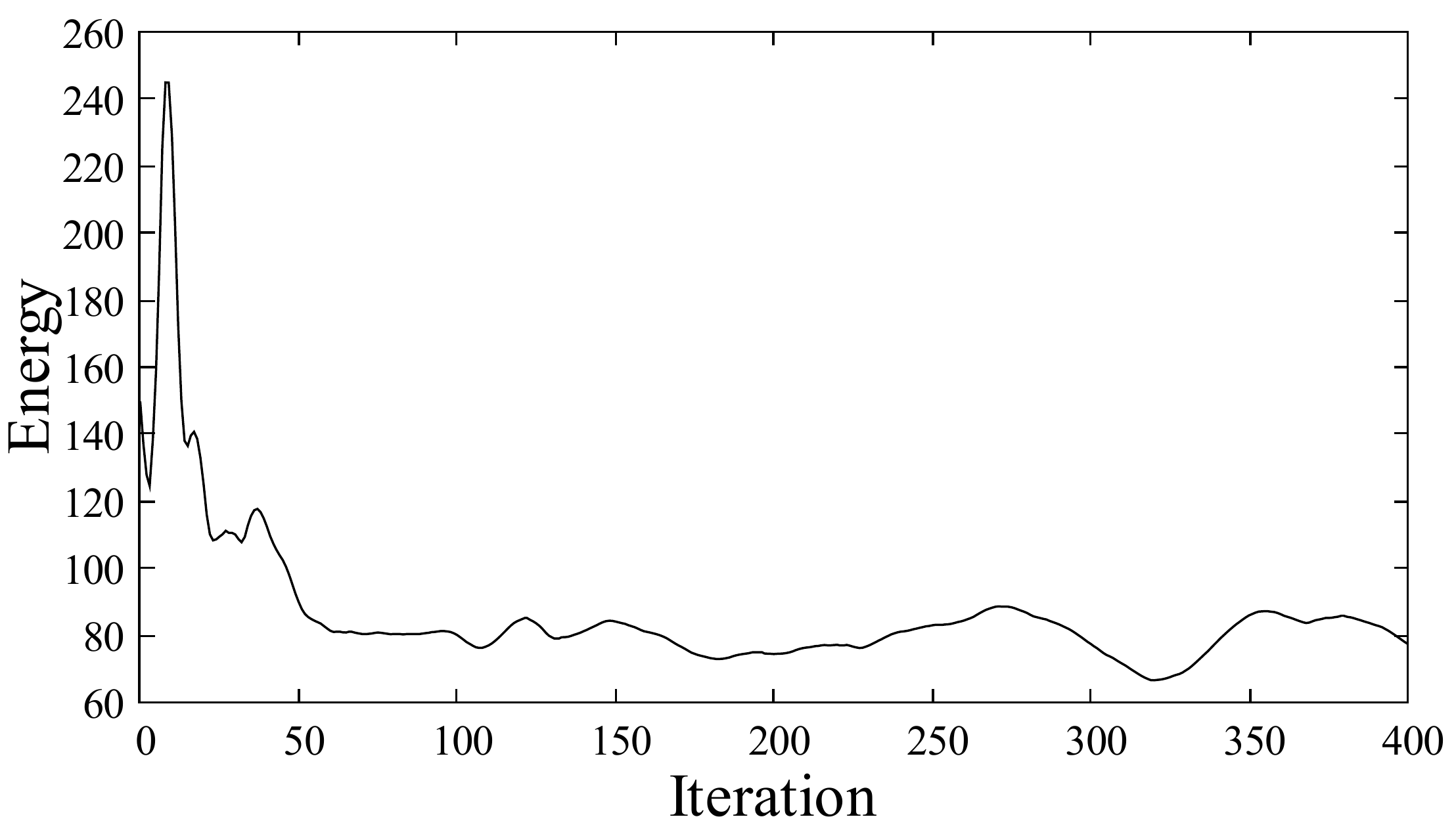}}
\hfil
\subfloat[Ours (IGNNS+B-ADT)]{\includegraphics[width=43mm]{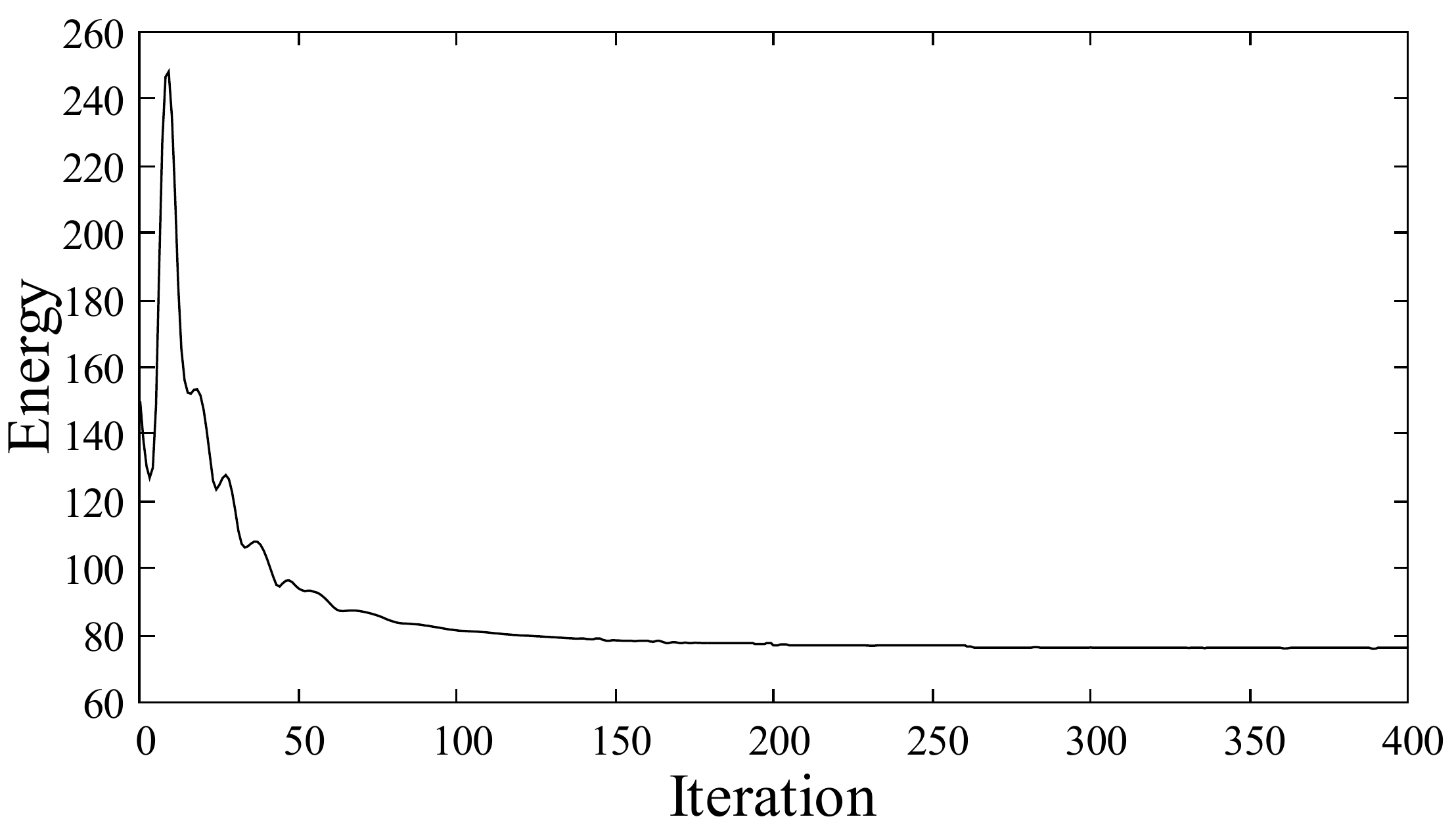}}
\caption{Convergence of the energy}
\label{fig_conv}
\end{figure}

We experimentally found that the energy minimization is difficult to converge stably with the sparse data term.
We compared the convergence of the energy minimization between our method (``IGNNS+B-ADT") and the baseline method (``B-ADT only").
Our method and the baseline method differ only in the data term of the energy.
The data term of ``B-ADT only" is based on the input sparse depth $d$.
We used the inverse of the depth as in our method.
Hence, the data term of baseline method $C^{\prime}(u)$ is defined as the following.
\begin{equation}
C^{\prime}(u) = \lambda_d w \left|  u - d^{-1} \right|^2
\end{equation}
With the same hyper parameters as in Section \ref{sec_baselines}, we observed the change in the energy during its minimization on Frame 12 of the Komaba dataset (Fig. \ref{fig_conv}).
We can see that the energy of our method is converging stably in contrast to the energy of ``B-ADT only" which oscillates through the iteration.
Note that the two methods have different energy and there is no point in comparing the values of the energy.

\section{Datasets details}
\label{app_data}
In this appendix, we explain the datasets introduced in Section \ref{sec_data} in more detail.
Note KITTI in this section is different from the benchmark dataset in Section \ref{sec_benchmark}.

Komaba is used in \cite{hirata2019real} and publicly available\footnote{https://github.com/menandro/upsampling}.
There are 56 image pairs with resolutions of 1238 $\times$ 374, dense depth maps captured with Faro Focus S150\footnote{https://www.faro.com/}, simulated LIDAR data, semantic segmentation results, motion stereo, and 7 manual segmentation labels.
We selected 5 frames (Frame 8, 12, 18, 29, 35) which are the same as frames used in the evaluation of \cite{hirata2019real}.
And we sampled dense depth maps with specifications of Velodyne VLP-16\footnote{https://velodyneLIDAR.com/}: vertical resolution 1.3 degree, horizontal resolution 0.2 degree, receptive range from 0.5 - 100 [m].

KITTI is introduced by \cite{geiger2013vision} and widely used.
We used manual semantic labels from Xu et al. \cite{xu2016multimodal}.
And we collected corresponding dense depth maps from the depth completion benchmark dataset \cite{uhrig2017sparsity}.
In this way, we collected 97 frames of dataset composed of scenes of 72 City, 14 Residential, 4 Campus, and 7 Road.

In KITTI, projected LIDAR depth maps contain the depth of the occluded background because of the large disparity between LIDAR and cameras.
We removed them by a simple pre-processing.
For every measured depth $d$, we draw a lower semi-circle with radius $r_{\rm occ}$ centered at the position of $d$, and remove depth if it is covered by the semi-circle and its value is farther than threshold $t_{\rm occ}$ comparing with the current semi-circle center.
Our pre-processing is based on the observation that occluded background depth locates at below of foreground object depth on the image plane.

\end{document}